\documentclass[preprint,12pt]{elsarticle}
\biboptions{sort&compress}
\usepackage{epstopdf}
\usepackage{rotating}
\usepackage{subfigure}
\usepackage{stfloats}
\usepackage{multirow}
\usepackage{color}
\usepackage{makecell}
\usepackage[dvipsnames]{xcolor}
\usepackage[table]{xcolor}
\usepackage{graphicx}
\usepackage{hyperref}
\usepackage{empheq}
\usepackage{latexsym}
\usepackage{placeins}
\usepackage{amssymb}
\usepackage{amsmath}
\usepackage{pifont}

\definecolor{mygreen}{RGB}{34,139,34}

\journal{Nuclear Physics B}

\begin{document}

\begin{frontmatter}




\title{EF-LLM: A Continual Learning-Based Framework for AI-Assisted Automation, Enhanced Sparse Prediction, and Hallucination Detection}


\author[1]{Zihang~Qiu}
\author[1]{Jayashri~Ravishankar}
\author[1]{Anam~Malik}
\author[2]{Zhongju~Wang}
\author[1]{Renyou~Xie}

\affiliation[1]{organization={school of EE\&T, UNSW Sydney},
            city={Sydney},
            postcode={2052}, 
            state={NSW},
            country={Australia}}

\affiliation[2]{organization={school of systems and computing, UNSW Canberra},
            city={Canberra},
            postcode={2610}, 
            state={ACT},
            country={Australia}}


\begin{abstract}
Accurate prediction helps achieve supply-demand balance in energy systems, thereby supporting operational decision-making and scheduling. 
Traditional models, which lack AI-assisted automation, often rely heavily on expert knowledge, incur high labor costs, and perform poorly under sparse-data conditions. To address these challenges, this work proposes an Energy Forecasting Large Language Model (EF-LLM), which integrates domain knowledge and temporal data for time-series forecasting while supporting both pre-forecast operations and post-forecast decision-making. 
EF-LLM’s human–artificial intelligence (AI) interaction capabilities further lower the entry barrier for forecasting tasks and reduce the need for additional expert involvement.
To support these functions, a continual learning framework with updatable Low-Rank Adaptation (LoRA) and a multi-channel architecture is developed to align heterogeneous multimodal data, enabling EF-LLM to continuously acquire and adapt multimodal knowledge.
In addition, EF-LLM improves prediction accuracy under sparse-data conditions by leveraging its ability to process and fuse multimodal information. 
A Fusion Parameter-Efficient Fine-Tuning (F-PEFT) method is further introduced to effectively exploit both time-series data and textual information.
EF-LLM is also the first energy-specific LLM to detect hallucinations and quantify their occurrence rate, achieved via multi-task learning and semantic similarity analysis.
This work has achieved success in energy prediction scenarios for load, PV, and wind power forecast.
\end{abstract}




\end{frontmatter}



\section{Introduction}   \label{SEC_introduction}
Forecasting is fundamental to energy systems, supporting tasks like power prediction, frequency regulation, etc. 
Current models mainly fall into two types: mathematical \& AI-based. 
Mathematical models (e.g., linear regression, Kalman filters) offer strong interpretability due to clear theoretical foundations \cite{linardatos2020explainable, lakkaraju2016interpretable, xing2015participation}. 
In contrast, AI models (e.g., RNNs, GANs, STGNNs) excel in handling nonlinear data \cite{abiodun2018state, nickel2012factorizing, barrett2013applying}.

However, both approaches have drawbacks. 
Traditional models, whether mathematical or AI-based, lack human-AI interaction for task processing automation, requiring expert intervention and increasing reliance on specialists, thus raising costs. 

Additionally, traditional models struggle with data sparsity. 
In this work, data sparsity refers to two common situations in energy forecasting. The first is feature category sparsity, where the available input features are limited and cannot provide sufficiently comprehensive information for the forecasting task. For example, only a small number of meteorological or operational variables may be available for load, photovoltaic, or wind forecasting, which restricts the representation ability of conventional forecasting models. The second is rare-event sparsity within features, where certain influential feature states or transition patterns occur only a few times in historical data. Typical examples include abrupt weather transitions such as “heavy rain turning to clear,” “clear turning to heavy rain,” “haze,” and “cloudy turning to overcast.” Although these events are rare, they may significantly affect energy output and therefore are highly important for forecasting. 
However, their extremely limited occurrences make them difficult for traditional models to learn and generalize reliably.

Large Language Models (LLMs) have the potential to address these challenges effectively. 
For AI-assisted automation, fine-tuning relevant knowledge enables the LLM to engage in interactive dialogue and operational guidance.
This reduces the need for expert input, allowing non-experts to manage forecasting tasks through simple LLM interactions.
For data sparsity, natural language descriptions offer flexibility, and LLMs -- trained on vast datasets -- leverage extensive knowledge bases to generalize the impact of sparse features and datasets. 
Fine-tuning with scenario-specific data further enhances their ability to quantify these effects, providing strong reasoning under data sparsity \cite{huang2024fewermoreboostingllm, li2024flexkbqa}. 

Current work on LLMs has yet to fully achieve the listed capabilities in energy forecasting field.
For AI-assisted forecasting automation and operation support (e.g., feature engineering \& decision support), existing LLM-based time-series forecasting models \cite{das2023decoder, liu2024autotimes, liu2024calfaligningllmstime, jin2023time} lack conversational capabilities and are limited in scalability.
Even in energy systems, these models' lack of conversational interaction limits their ability to handle complete forecasting tasks, hindering non-experts from using LLMs for AI-assisted automation.
Though studies like \cite{wu2024stellm, xue2023promptcast} highlight this potential, they lack structures to enable such functionalities. 
Implementing these would allow LLMs to manage the entire forecasting process -- data input, feature engineering, prediction, and post-decision support -- reducing reliance on experts, lowering entry barriers, and cutting costs for specialized expertise.

Additionally, some researchers have addressed sparse data prediction using LLMs.
In \cite{shi2024language}, LLMs leverage pre-trained knowledge to cluster sparse events, thereby enhancing prediction in such scenarios. 
Similarly, \cite{li2024cllmate} integrates language descriptions into LLMs to enrich image data, improving weather event predictions.
Although LLMs have successfully addressed data sparsity issues in other domains, these studies have not been applied to energy systems and lack the capability to handle time-series forecasting.
Additional limitations exist in time-series LLMs using prefix tuning -- a common fine-tuning approach \cite{li2021prefix} -- to embed multimodal data for improved predictions \cite{das2023decoder, liu2024autotimes, liu2024calfaligningllmstime, jin2023time, wu2024stellm}. 
However, \cite{liu2024autotimes} incorporates text only for representing timestamps, while \cite{das2023decoder, liu2024calfaligningllmstime, wu2024stellm} simply repeat the time-series content in textual number. 
In \cite{jin2023time}, text-based feature engineering is limited to basic stats like max, min, and mean.
These methods miss leveraging LLMs' unique textual processing capabilities.
As a result, the text inputs could be replaced with numerical features, rendering the text redundant.
A primary reason for using textual inputs is to activate the pre-trained knowledge in LLMs, leveraging relevant information from the LLM's knowledge base to enhance prediction performance \cite{lian2023llm, lee2024llm2llm, wang2024enhancing}. 
However, this potential remains unrealized in the aforementioned multimodal LLM studies \cite{das2023decoder, liu2024autotimes, liu2024calfaligningllmstime, jin2023time, wu2024stellm}.

Furthermore, leveraging LLMs to reduce barriers, automate tasks, and cut labor costs has become an important trend in the energy field \cite{wu2024stellm, zhang2024large, wang2024power, zhou2024elecbench, jiang2024eplus}. Specialized LLMs with human-AI interaction capabilities have been developed for a variety of task scenarios within energy systems, such as EPlus-LLM for automated load data trend modeling \cite{jiang2024eplus}, STELLM for wind speed forecasting \cite{wu2024stellm}, POWER-LLAVA for identifying transmission line risks \cite{wang2024power}, LLM-assisted BESS optimization in FCAS markets \cite{zhang2024large}, and ElecBench for power dispatch evaluation \cite{zhou2024elecbench}. These studies demonstrate that LLMs can achieve solid performance across diverse energy-related tasks and highlight their potential to support AI-assisted forecasting and decision-making.

However, hallucination remains a critical reliability issue for LLMs in energy forecasting and related operational tasks \cite{rawte2023survey, li2024banishing, farquhar2024detecting, zhang2023siren, tonmoy2024comprehensive}. 
{Hallucination is usually defined as fabricated explanations, logically inconsistent statements, or outputs that deviate from the expected response structure in textual responses.}
Such errors can undermine the trustworthiness of AI-assisted forecasting and decision support, hallucination detection is essential for evaluating whether an energy-oriented LLM is reliable enough for practical deployment. 
Nevertheless, despite the promising performance of existing energy-domain LLMs, none of the aforementioned studies explicitly address or propose methods for detecting hallucinations, leaving an important gap in the reliability assessment of domain-specific LLMs.

To solve the challenges, there is a central question: 
How to train an energy-specific LLM that offers complete process forecasting service through human-AI interaction, enhances prediction accuracy under data sparsity, and transparently detects hallucinations?
To address this, this work develops a foundational Energy Forecasting LLM (EF-LLM).
The fine-tuned EF-LLM offers the following contributions:

1.
AI-assisted automation: 
EF-LLM integrates the full predictive logic chain for photovoltaic (PV), load, and wind power forecasting, spanning operational guidance, feature engineering, prediction, and post-decision support via user interaction.
This streamlines forecasting as EF-LLM handles most tasks, needing only user data and prompts, reducing reliance on skilled engineers and cutting costs.
This work integrates heterogeneous multimodal knowledge through a multi-channel architecture and adopts a LoRA-based continual learning approach to continuously update knowledge, enabling EF-LLM to efficiently handle diverse tasks.

2.
LLM-based sparse prediction:
EF-LLM excels in predictive accuracy by extracting insights from multimodal data, including time-series and text.
{{Through a continual learning strategy, EF-LLM sequentially incorporates textual and time-series information from both sparse features and complete data, enabling the model to maintain sensitivity to sparse features while remaining robust to sparse data.
During inference, through sequential textual injection based on chain of thought(CoT), textual descriptions can improve prediction accuracy under data-sparse conditions. 
The proposed Fusion Parameter-Efficient Fine-Tuning (F-PEFT) method combines prefix and LoRA, enabling EF-LLM to effectively utilize both types of data.}
During inference, through sequential textual injection based on chain of thought(CoT), textual descriptions can improve prediction accuracy under data-sparse conditions. 
The proposed Fusion Parameter-Efficient Fine-Tuning (F-PEFT) method combines prefix and LoRA, enabling EF-LLM to effectively utilize both types of data.}

3.
Hallucination Analysis:
{EF-LLM employs multi-task learning and semantic similarity analysis to quantify the impact of task importance on the hallucination rate in textual responses. 
To the best of our knowledge, EF-LLM is the first LLM in the energy field that can detect and quantitatively demonstrate hallucination rates, reflecting its strong reliability in energy system tasks.}

The rest of paper is structures as:
Section \ref{SEC_related} highlights the efficacy of EF-LLMs with proven methods in the literature.
Section \ref{SEC_framework} introduces  the detailed architecture of EF-LLM and the methodological details adopted in this paper. 
Section \ref{SEC_automation} corresponds to the first contribution, AI-assisted automation, showcasing the architecture and techniques that enable EF-LLM to learn new knowledge in energy forecasting domain and gain function-calling capabilities.
Section \ref{SEC_sparse} presents F-PEFT, corresponding to the second contribution, LLM-based sparse prediction.
{Section \ref{SEC_hallucination} explains the methods for quantitatively analyzing hallucination, addressing the third contribution.
Section \ref{SEC_reliability} explains the ANOVA method used to statistically analyze the reliability of the numerical output component of EF-LLM.}
Finally, section \ref{SEC_CaseStudy} shows a case study, \ref{SEC_future_work} the future work and finally \ref{SEC_conclusion} the conclusion.

\section{Related Work} \label{SEC_related}

To provide a more intuitive comparison between existing energy-domain LLMs and EF-LLM in energy forecasting tasks, a systematic summary is presented in Table \ref{Tab_Energy_LLM_Review}.

  \begin{table} [!b]
    \centering
    \caption{{Capability Summary of Energy-domain LLMs}}
        \label{Tab_Energy_LLM_Review}
    \renewcommand\arraystretch{0.9}
    	\setlength{\tabcolsep}{1.5mm}{
    \begin{tabular} {c|c|c|c|c|c|c}
        \hline
        \hline
                \textbf{Func}& \textbf{Auto} &  \textbf{Pred}  &  \textbf{Multimodal}  &  \makecell{\textbf{Hallucination} \\[-2pt] \textbf{Detection}}  &  \makecell{\textbf{multiple} \\[-2pt] \textbf{scenarios}} & \textbf{scope} \\
        \hline
        \cite{zhang2024large} &  \textcolor{mygreen}{\ding{52}}  &  \textcolor{red}{\ding{56}}  &  \textcolor{red}{\ding{56}}  &  \textcolor{red}{\ding{56}}  &  \textcolor{red}{\ding{56}} & bidding \\
                \hline
        \cite{liang2025energygpt} &  \textcolor{mygreen}{\ding{52}}  &  \textcolor{red}{\ding{56}}  &  \textcolor{red}{\ding{56}}  &  \textcolor{red}{\ding{56}}  &  \textcolor{red}{\ding{56}} & prediction \\
        \hline
        \cite{wang2024power} &  \textcolor{red}{\ding{56}}  &   \textcolor{red}{\ding{56}}  &  \textcolor{mygreen}{\ding{52}}  &   \textcolor{red}{\ding{56}}  &  \textcolor{red}{\ding{56}} & Knowledge QA \\
        \hline
        \cite{bao2026energy} &  \textcolor{red}{\ding{56}}  &  \textcolor{red}{\ding{56}}  &  \textcolor{red}{\ding{56}}  &   \textcolor{mygreen}{\ding{52}}  &  \textcolor{red}{\ding{56}} & Knowledge QA \\
        \hline
        \cite{gabber2024domain} &  \textcolor{red}{\ding{56}}  &  \textcolor{red}{\ding{56}}  &  \textcolor{red}{\ding{56}}  &  \textcolor{red}{\ding{56}}  &  \textcolor{mygreen}{\ding{52}} & Knowledge QA  \\
        \hline
        \cite{buster2023large} &  \textcolor{red}{\ding{56}}  &  \textcolor{red}{\ding{56}}  &  \textcolor{red}{\ding{56}}  &  \textcolor{red}{\ding{56}}  &  \textcolor{mygreen}{\ding{52}} & Knowledge QA  \\
        \hline
        \cite{zhang2025elecbench} &  \textcolor{red}{\ding{56}}  &  \textcolor{red}{\ding{56}}  &  \textcolor{red}{\ding{56}}  &  $\blacktriangle$  &  \textbf{\textcolor{mygreen}{\ding{52}}} & dispatch \\
        \hline
        \cite{jiang2024eplus} &  \textcolor{mygreen}{\ding{52}}  &  \textcolor{red}{\ding{56}}  &  \textcolor{mygreen}{\ding{52}}  &  \textcolor{red}{\ding{56}}  &  \textcolor{red}{\ding{56}} & modelling \\
        \hline
        \cite{cheng2025large} &  \textcolor{red}{\ding{56}}  &  $\blacktriangle$  &  \textcolor{mygreen}{\ding{52}}  &  $\blacktriangle$  &  \textcolor{red}{\ding{56}} & dispatch \\
        \hline
        \cite{wu2024stellm} &  \textcolor{mygreen}{\ding{52}}  &  \textcolor{mygreen}{\ding{52}}  &  \textcolor{mygreen}{\ding{52}}   & \textcolor{red}{\ding{56}}  &  \textcolor{red}{\ding{56}} & prediction \\
        \hline
        \rowcolor{gray!10}
        EF-LLM &  \textcolor{mygreen}{\ding{52}}  &  \textcolor{mygreen}{\ding{52}}  &  \textcolor{mygreen}{\ding{52}}  &  \textcolor{mygreen}{\ding{52}}  &  \textcolor{mygreen}{\ding{52}} & prediction \\
        \hline
        \hline
    \end{tabular}  }
    \vspace{-0.5em}
    \begin{minipage}{0.95\linewidth}
        \scriptsize
        $\blacktriangle$: potentially capable;  \textcolor{mygreen}{\ding{52}}: capable; \textcolor{red}{\ding{56}}: unable.    
    \end{minipage}
  \end{table}

EF-LLM establishes a relatively complete and reliable framework for energy forecasting, characterized by four key capabilities. 
First, automation, meaning that it reduces the dependence on human intervention throughout the forecasting workflow and enables a higher degree of process automation. Second, prediction, meaning that EF-LLM can process time-series data and perform forecasting using its own internal architecture, without relying on external forecasting models. 
Third, multimodal, meaning that the model itself is capable of multimodal information injection and processing, rather than depending on external tools to separately handle different modalities before combining them. Fourth, hallucination detection, meaning that it can identify hallucination phenomena in the generated outputs and provide corresponding indicators to evaluate the occurrence rate of hallucinations or the degree of output unreliability.

In Table \ref{Tab_Energy_LLM_Review}, Func, Auto, and Pred are abbreviations for function, automation and prediction, respectively.

As shown in Table \ref{Tab_Energy_LLM_Review}, most existing energy-domain LLMs still exhibit various limitations in achieving automated energy forecasting. 
In contrast, EF-LLM not only enables automated workflow execution, but also performs forecasting and multimodal information injection through its own architecture. This characteristic allows EF-LLM to support multi-stage textual information supplementation and interactive forecasting, thereby showing stronger potential under sparse-data conditions, which is a capability that most existing energy-domain LLMs do not yet systematically possess.

In addition, except for one study that have explicitly investigated hallucination detection \cite{bao2026energy}, most existing models still lack a systematic and explicit analysis of hallucination rates. As a result, the reliability of their outputs remains somewhat uncertain in practical applications.

Overall, EF-LLM simultaneously integrates the four key capabilities of automation, prediction, multimodal information injection, and hallucination detection, making it a highly representative and methodologically significant framework for energy forecasting.

{It should be noted that due to differences in the scope and objectives of the studies, it is difficult to conduct a strictly fair quantitative performance comparison across all these models. Therefore, the comparison in this work is mainly conducted among prediction-oriented methods under a consistent forecasting setting \cite{liang2025energygpt, wu2024stellm}.}

{In particular, STELLM is more closely aligned with the design philosophy of EF-LLM, as both aim to incorporate as much domain knowledge as possible into a single standalone LLM for energy forecasting tasks. From this perspective, STELLM is the most appropriate baseline for quantitative comparison.}

\section{EF-LLM Complete Technical Framework}   \label{SEC_framework}

  \begin{figure} [!htb]
	\centering
	\includegraphics[width=5.5in]{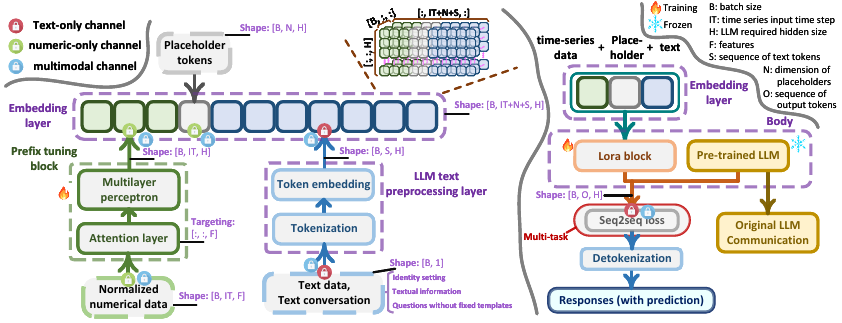}
	\caption{EF-LLM Complete Framework.}
	\label{framework_all}
  \end{figure}

Fig.\ref{framework_all} presents the complete technical framework composed of all modules in EF-LLM. The principles and descriptions of each module are presented in the following subsections.

{In the framework, there are two channels: a text-only channel, and a multimodal channel.}

\subsection{Framework of Each Module}\label{SEC_module}
\subsubsection{EF-LLM Base Framework}
\begin{figure} [!htb]
	\centering
	\includegraphics[width=5.5in]{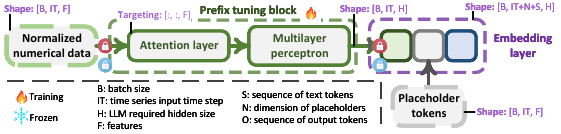}
	\caption{EF-LLM Prefix Tuning Block.}
	\label{framework_1}
  \end{figure}

\begin{figure} [!htb]
	\centering
	\includegraphics[width=5.5in]{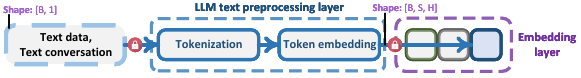}
	\caption{EF-LLM text preprocessing layer.}
	\label{framework_2}
  \end{figure}

\begin{figure} [!htb]
	\centering
	\includegraphics[width=5.5in]{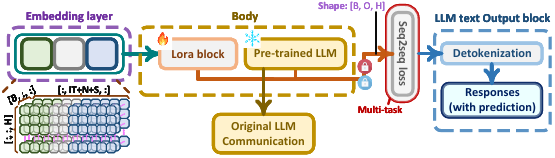}
	\caption{EF-LLM Deep Network Architecture with LoRA.}
	\label{framework_3}
  \end{figure}

EF-LLM Base consists of three parts, as shown in Fig.\ref{framework_1}, \ref{framework_2}, \ref{framework_3}. 
Fig.\ref{framework_1} illustrates the meaning of each letter and symbol.
EF-LLM framework includes the following components:

\textbf{Prefix Tuning Block:} This block is used to handle pure time series information, as shown in Fig.\ref{framework_1}.

The input time-series data are normalized. 
The prefix tuning block can carry pre-trained time-series information if required \cite{jin2023time, chow2024towards}. 
This block can process and transmit temporal information, which plays a guiding role in word associations within the high-dimensional vector space of EF-LLM.

In this module, the attention mechanism is adopted:
\begin{equation}\label{S2_1}
\alpha(Q, K, V) = softmax(\frac{QK^T}{\sqrt{d_k}})V,
\end{equation}
where $Q$, $K$, $V$ are query matrix, key matrix, and value matrix respectively.
$\alpha$ is the attention output.
\begin{subequations}\label{S2_2}
\begin{align}
&h^N_1 = ReLU(W_1 \cdot \alpha + b_1), \label{S2_2a} \\
&h^N_n = ReLU(W_n \cdot h_{n-1} + b_n). \label{S2_2b}
\end{align}
\end{subequations}

Eq.\eqref{S2_2} represents a multilinear perceptron (MLP) connected after the attention output, where $h^N_n$ and $b_n$ are the output and bias of the $n$-th layer of the MLP, respectively.
The ReLU activation function is used to eliminate negative values.

\textbf{LLM Text Preprocessing Layer:} The input textual information, also known as the textual prompt, is mapped to the high-dimensional vector space of EF-LLM's pretrained knowledge base through its tokenizer, as shown in Fig.\ref{framework_2}.

As in Eq.\eqref{S2_3a}, the textual information is first tokenized by the LLM's tokenizer into a sequence of words, where $m^i$ is the number of input words.
Then, as in Eq.\eqref{S2_3b}, each word $w_i$ is mapped to a high-dimensional vector space $v_i$ through the embedding matrix $E$, where $E$ is a $V \times d$ matrix, $V$ is the size of the LLM's vocabulary, and $d$ is the dimension of the embedding vectors.
Ultimately, as in Eq.\eqref{S2_3c}, these high-dimensional vector sequences form the textual information $h^T$.
\begin{subequations}\label{S2_3}
\begin{align}
&\{w_1, w_2, \cdots, w_{m^i}\} = \text{Tokenizer}(\text{text}), \label{S2_3a} \\
&v_i = E(w_i), \label{S2_3b} \\
&h^T = \{v_1, v_2, \cdots, v_{m^i}\}. \label{S2_3c}
\end{align}
\end{subequations}

\textbf{Embedding Layer:} As shown in Fig.\ref{framework_1}, this layer comprises numerical input, placeholder tokens, and text tokens. 
Placeholder tokens are used to separate numerical input and text tokens dimensionally, as in Eq.\eqref{S2_4}.
\begin{subequations}\label{S2_4}
\begin{align}
&h^P = \text{InvalidChar}, \label{S2_4a} \\
&h^E = \text{SeqEmbedding}[h^N: h^P: h^T]. \label{S2_4b}
\end{align}
\end{subequations}

$h^P$ is placeholder sequence.
$h^E$ is the concatenated input.

\textbf{Body with LoRA:} LoRA is used to restructure word associations.
With LoRA activated, EF-LLM can handle specific tasks in energy forecasting domain.
When LoRA is not activated, the pre-trained base model is used, as shown in Fig.\ref{framework_3}, Eq.\eqref{S2_5}.

In Eq.\eqref{S2_5}, $O^{LLM}$ is the output of LLM deep neural network.
\begin{equation}\label{S2_5}
O^{LLM} = \text{LLM \& LoRA}(h^E).
\end{equation}

LoRA adds a low-rank matrix to the pre-trained LLM base, as in Eq.\eqref{S1_1}, $W_0$ is pre-trained LLM's weights, $\Delta W$ is the weights of the LoRA block on the LLM, $W$ is the model weight after fine-tuning. 
$A$ and $B$ are two low-rank matrices to reduce the number of parameters in $\Delta W$ while training. 
\begin{equation}\label{S1_1}
W = W_0 + \Delta W = W_0 + AB^T.
\end{equation}

\textbf{Output Part:} After the LLM processes the tensors, it directly detokenizes the tensors to achieve textual output.
\begin{subequations}\label{S2_6}
\begin{align}
&O^{LLM} = \{v^O_1, v^O_2, \cdots, v^O_{m^o}\}, \label{S2_6a} \\
&w^O_i = \text{Detokenizer}(v^O_i), \label{S2_6b} \\
&\text{text} = \text{Concatenate} (w^O_1, w^O_2, \cdots, w^O_{m^o}). \label{S2_6c}
\end{align}
\end{subequations}

Eq.\eqref{S2_6} is detokenization process, 
$m^o$ is output word number.

\subsubsection{EF-LLM Multi-task Loss function}
EF-LLM's AI-assisted capability is based on human-AI interaction through F-PEFT, so this work also uses Seq2Seq loss like the pre-trained LLM base.
While training EF-LLM, prefix tuning does not change the loss function of the pre-trained LLM, but LoRA does.
Additionally, this work adopts a multi-task learning framework, which is used for sensitivity analysis, helping to quantify and demonstrate how the hallucination rate varies with changes in the multi-task coefficient.
EF-LLM's final loss function is shown as Eq.\eqref{S1_2}.
\begin{subequations}\label{S1_2}
\begin{align}
&\mathcal{L}_{1} = - \sum_{k=1}^{m_1} \log P(y_k^1 | \mathbf{Y}^1_{1:k-1}, \mathbf{X}^1), \label{S1_2a} \\
&\mathcal{L}_{2} = - \sum_{k=1}^{m_2} \log P(y_k^2 | \mathbf{Y}^2_{1:k-1}, \mathbf{X}^2), \label{S1_2b} \\
&\mathcal{L} = \mathcal{L}_{\text{orig}} + \lambda \|\Delta W\|_F^2, \label{S1_2c} \\
&\mathcal{L}_{\text{orig}} = \varpi \cdot \mathcal{L}_{1} + (1-\varpi) \cdot \mathcal{L}_{2}, \label{S1_2d} \\
&m = m_1 + m_2.  \label{S1_2e}
\end{align}
\end{subequations}

Eq.\eqref{S1_2a}\eqref{S1_2b} represent Tasks $\mathcal{L}_{1}$ and $\mathcal{L}_{2}$, where Seq2Seq loss is to use the preceding token sequence ${Y}_{1:k-1}$ and the input $X$ to accurately predict the target word $y_k$ at the $k$-th token position.
In Eq.\eqref{S1_2c}, $\|\Delta W\|_F$ is the Frobenius norm of $\Delta W$, and $\lambda$ is a regularization parameter.
In Eq.\eqref{S1_2d}, $\mathcal{L}_{\text{orig}}$ is the original loss function of the pretrained LLM base.
$m$ represents the length of the token sequence output by the LLM.
$\varpi$ represents the relative importance of Task $\mathcal{L}_{1}$ and Task $\mathcal{L}_{2}$. 
$m_1$ and $m_2$ denote the token ranges corresponding to Task $\mathcal{L}_{1}$ and Task $\mathcal{L}_{2}$, respectively, when using multi-task learning.
In Eq.\eqref{S1_2e}, $m$ is the token range of the whole tasks.
$P(y_k | \mathbf{Y}_{1:k-1}, \mathbf{X})$ denotes the probability that EF-LLM correctly outputs $y_k$.

It is worth mentioning that the loss function of LLMs makes them difficult to quantify the differences between numerical values; for instance, the loss between 3.1 and 3.2 is treated the same as the loss between 1 and 2, even though the difference between 1 and 2 is significantly greater. 
This limitation constrains the predictive ability of LLMs in regression tasks \cite{chow2024towards} and makes them more suited for classification tasks.
Therefore, a multi-task learning approach is adopted, separately presenting the accuracy for both classification tasks and regression tasks.
Task 1 involves interval prediction, which is a textual classification problem.
Task 2 is a textual regression problem.

\begin{equation}\label{S_1}
f(P_t) = \left\{
\begin{array}{lcr}
0, \quad \text{if} \quad P_t = 0; \\
i, \quad \text{if} \quad \frac{i-1}{N} \cdot E_r <  P_t \leq \frac{i}{N}, \quad \text{for} \quad i = 1, \cdots N.
\end{array}
\right.
\end{equation}

As an example, Eq.\ref{S_1} represents the classification method for the PV classification task in this paper.
$P_t$ is PV power generation at time $t$, $E_r$ is the rated grid-connected capacity, and $N$ is the number of intervals into which $E_r$ is divided.
Typically, the rated capacity of 798 kW is divided into 100 intervals, plus an additional interval for when $P_t = 0$, resulting in a total of 101 classes: $\{0, 0\sim1\%, 1\sim2\%, \cdots, 99\sim100\% \}$.

\subsubsection{EF-LLM With Updated LoRA}
As described above, EF-LLM consists of layers as in Eq.\eqref{Su_1}, where the highlighted layers in red represent layers with trainable weights, constituting $W$ in Eq.\eqref{S1_1}. 
LoRA, as an externally attachable trainable module, can be integrated with any weight layer of the original model.
By attaching an updatable LoRA $\Delta W'$ to the weight layers of the EF-LLM base $W$, freezing $W$, updating $\Delta W'$, EF-LLM's knowledge can be continuously updated to achieve continual learning.
\begin{equation}\label{Su_1}
\text{EF-LLM base} = 
\left\{
\begin{array}{lcr}
\textcolor{Red}{\text{Prefix Tuning Block}} \\
\text{Text Preprocessing Layer} \\
\text{Embedding Layer} \\
\textcolor{Red}{\text{Body With LoRA}} \\
\text{Output Layer} 
\end{array}
\right.
\end{equation}

Furthermore, as in Eq.\eqref{S1_2c}, LoRA finetuning includes a penalty term $\lambda \|\Delta W\|_F^2$, which ensures minimal alteration of the base model's knowledge. 
This allows continual learning based on updated LoRA to continuously enhance EF-LLM's knowledge while preserving the integrity of EF-LLM knowledge base.
\begin{equation}\label{Su_2}
W' = W + \Delta W' = W + A'B'^T.
\end{equation}

Eq.\eqref{Su_2} shows the externally updated LoRA structure $\Delta W'$ of EF-LLM, which differs from the LoRA $\Delta W$ in Eq.\eqref{S1_1}. 
Specifically, $\Delta W$ is attached only to the LLM base. 
After training the EF-LLM base, $\Delta W'$ is added to the EF-LLM base and incorporates the newly introduced prefix tuning block.

\subsection{AI-assisted Automation}   \label{SEC_automation}
EF-LLM's ability to handle each step of the complete forecasting process requires it to possess both time-series forecasting and human-AI interaction capabilities. 
To achieve this, this work adopts the following techniques: 
First, this work designs a multi-channel architecture and a heterogeneous multimodal data alignment technique, enabling simultaneous training of textual and time-series knowledge, ensuring various types of knowledge can be effectively learned by EF-LLM.
Moreover, training EF-LLM based on a pretrained LLM base requires significant time. 
To avoid redundant training and continuously integrate new knowledge, EF-LLM adopted a LoRA-based continual learning approach, enabling EF-LLM to update its knowledge effectively.
Finally, EF-LLM employs function calling, allowing it to handle prompt \& feature engineering, and complex mathematical computations that are challenging for neural networks.
This section introduces these key technologies and explains the principles behind their implementation.

\subsubsection{Heterogeneous Multimodal Data Alignment}
\begin{figure} [!htb]
	\centering
	\includegraphics[width=5.5in]{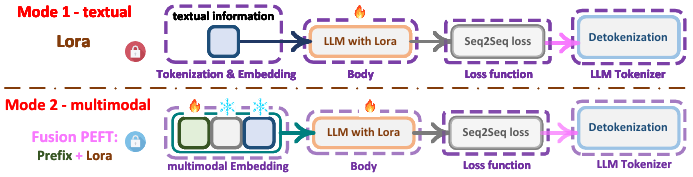}
	\caption{Heterogeneous Data Alignment: Multi-Channel Architecture.}
	\label{heterogenous_align}
  \end{figure}

The multi-channel architecture is as shown in Fig.\ref{heterogenous_align}.
Achieving human-AI interaction requires us to incorporate two types of knowledge into EF-LLM. 
The first type is forecasting-related knowledge, including power regulations and operational guidance, introduced in textual form, corresponding to Mode 1 in Fig.\ref{heterogenous_align}.
The second type is multimodal data that combines numerical time-series data with language data describing sparse feature and data, corresponding to Mode 2 in Fig.\ref{heterogenous_align}.
The pathways of the channels are represented by red and blue lock symbols, as also indicated in Fig\ref{framework_1}, \ref{framework_2}, \ref{framework_3}.

Since the two types of knowledge are heterogeneous, EF-LLM needs to ensure such heterogeneous data can be trained simultaneously during the finetuned training process of EF-LLM.

  \begin{figure} [!htb]
	\centering
	\includegraphics[width=5.5in]{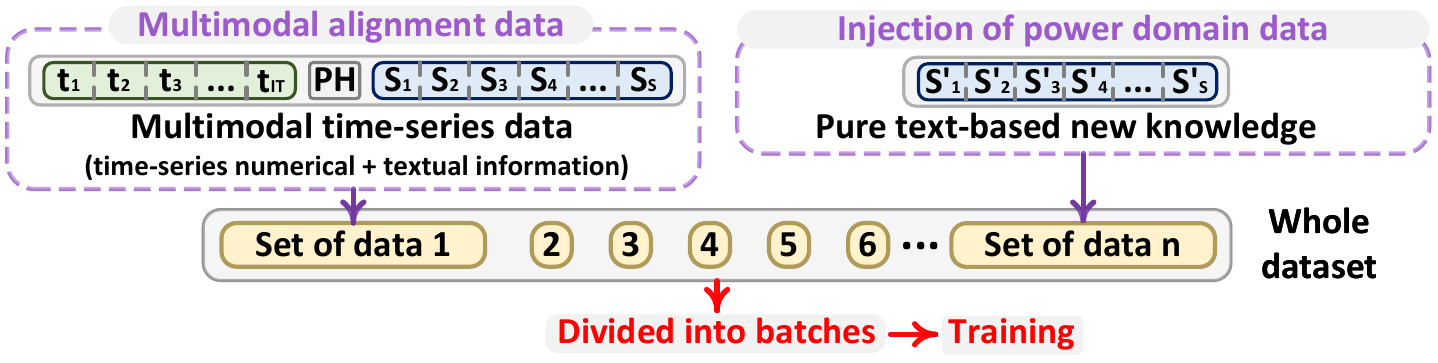}
	\caption{Heterogenous Multimodal Data Alignment.}
	\label{multimodal_align}
  \end{figure}

The multi-channel structure in Fig.\ref{heterogenous_align} enables heterogeneous data to be input into EF-LLM. Subsequently, EF-LLM uses the alignment method shown in Fig.\ref{multimodal_align} to inject the knowledge of this data into EF-LLM.
The data is arranged together in the dataset and injected into EF-LLM in multiple batches.

\subsubsection{Continuous Knowledge Injection}
\begin{figure} [!htb]
	\centering
	\includegraphics[width=5.5in]{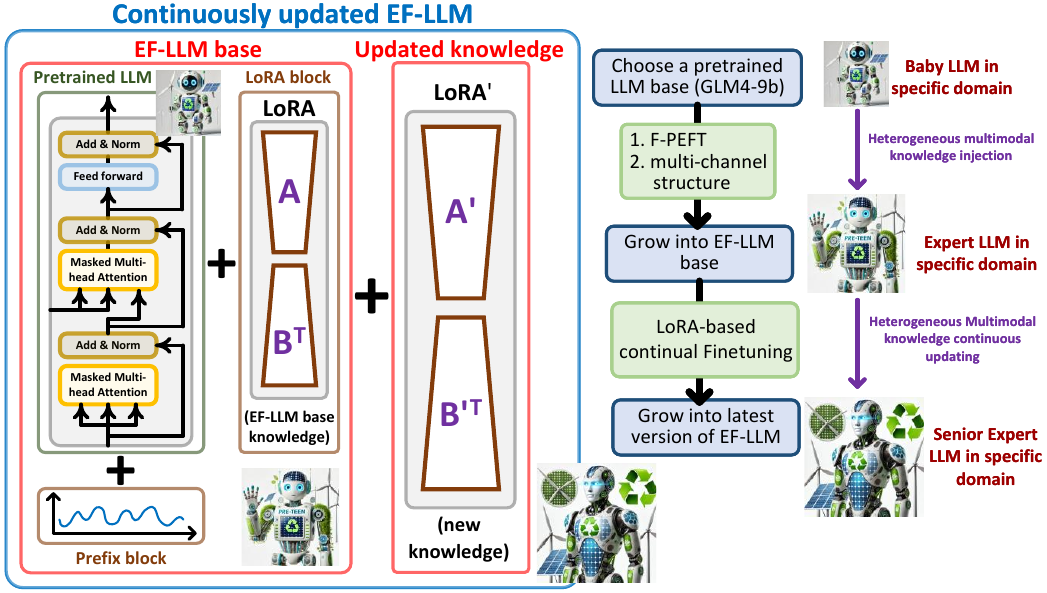}
	\caption{LoRA-based Continual Learning For EF-LLM.}
	\label{continual_learning_LoRA}
  \end{figure}

Finetuning EF-LLM consumes significant computational resources and time.
When new knowledge emerges, EF-LLM is updated through a LoRA-based continual learning method \cite{ren2024analyzingreducingcatastrophicforgetting}, as shown in Fig.\ref{continual_learning_LoRA}.

First, EF-LLM selects a pretrained LLM base, which possesses extensive natural language knowledge stored within its deep network. 
However, in the field of energy forecasting, it is merely a ``baby LLM" lacking the ability to respond with relevant knowledge.
Then, through F-PEFT and multi-channel structure, EF-LLM injects domain-specific knowledge into the pretrained LLM base, transforming it into the EF-LLM base. 
The current EF-LLM base becomes an expert in the field of energy forecasting and can independently provide prediction results without calling external functions. 
EF-LLM base's knowledge is stored within F-PEFT LoRA and Prefix blocks.

Over time, temporal drift may occur in time-series predictions, and textual knowledge may require updates. 
At this stage, EF-LLM uses updatable $LoRA'$. 
Without altering EF-LLM base, new temporal and textual data is injected to update its knowledge within $LoRA'$. 
This ensures that the updated EF-LLM consistently maintains its performance, ultimately becoming a senior expert LLM in energy forecasting domain.

\subsubsection{Function Calling}
\begin{figure} [!htb]
	\centering
	\includegraphics[width=5.5in]{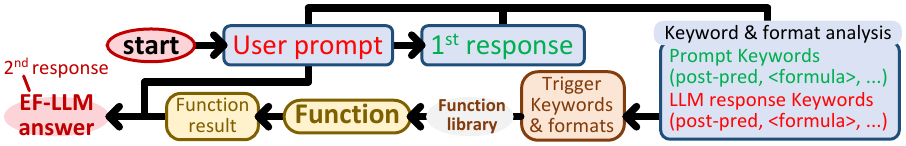}
	\caption{Function Call Example of EF-LLM.}
	\label{call_FPE_LLM}
  \end{figure}

The fine-tuning process of EF-LLM is essentially a gradient descent-based approach that enables the model to approximate the target. 
However, this approach has limitations in many tasks -- for instance, in tasks without standard target to approximate, such as prompt engineering and feature engineering; or in tasks requiring explicit solutions rather than mere approximations, such as complex mathematical calculations accurate to several decimal places. 
These limitations prevent EF-LLM from performing well on such tasks solely on its own.
This issue exists across various mainstream open-source LLMs. 
OpenAI was the first to provide a reliable solution to this problem: function-calling technology \cite{openai_function_calling_2023}, which is a core component of the LLM agents \cite{mei2024aios, talebirad2023multi, anthropicClaudeFunctionCalling, OpenAIFunctionCalling}.
To address these types of tasks, EF-LLM adopted function-calling technology. 
Fig.\ref{call_FPE_LLM} illustrates a case study of its application within EF-LLM.

First, based on user prompt, the first response is obtained. 
Then, the program analyzes the user prompt and EF-LLM response using a keyword $\&$ format retrieval method. 
EF-LLM will determine whether to activate a corresponding function in the function library. 
Once the conditions trigger the functions in the function library, the corresponding functions are executed. 
Next, the function results are used as part of the prompt, and together with the user prompt, sent to EF-LLM. 
The second response contains the mathematical calculation result from the function and is used as the final output.

\subsection{{LLM-based Sparse Prediction}}   \label{SEC_sparse}
{EF-LLM enhances sparse information through a targeted continual learning process. 
First, the LLM is supervised fine-tuned (SFT) on sparse data represented in text-only form, enabling it to recognize and retain sparse features. 
This stage encourages the LLM to prioritize sparse information and strengthens the latent knowledge in the pre-trained LLM that is related to the influence of sparse factors. 
Subsequently, EF-LLM performs F-PEFT on multimodal data composed of time-series data and textual data. 
This process follows the continual learning mechanism illustrated in Fig.\ref{continual_learning_LoRA} and reflects the design rationale of first reinforcing sparse-information modeling and then generalizing to the overall data distribution. Through this process, the textual modality can amplify and express knowledge related to sparse information, forming a complementary relationship with the numerical information in the time-series modality, thereby improving the predictive performance of EF-LLM under data sparsity.}

\subsubsection{{Design Rationale}} 
EF-LLM uses multimodal data inputs, leveraging the numerical modality to perform time-series predictions, while employing the textual modality to represent sparse features and enhance sparse data.
This enables EF-LLM to achieve better predictive performance under data sparsity. 
As demonstrated in Section \ref{SEC_framework}, EF-LLM proposes F-PEFT in EF-LLM, which integrates Prefix Tuning and LoRA Fine-tuning.

  \begin{figure} [!htb]
	\centering
	\includegraphics[width=5.5in]{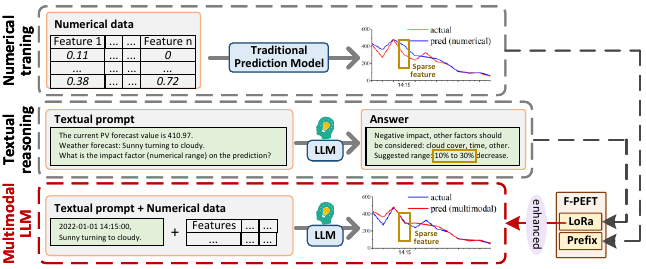}
	\caption{How F-PEFT Enhance Sparse Prediction.}
	\label{How_enhance_sparse}
  \end{figure}

The reason for this approach lies in the characteristics of the two fine-tuning methods, Prefix and LoRA. 
Prefix adds additional trainable modules before LLM deep network, guiding the LLM to extract more relevant knowledge during fine-tuning and enabling domain-specific adaptation. 
This method is task-specific and assumes the LLM already possesses relevant knowledge of the target task \cite{li2021prefix, jin2023time}. 
On the other hand, LoRA achieves fine-tuning by adding equivalent low-rank matrices to the deep network of the LLM, reconstructing word associations in the high-dimensional vector space and injecting domain-specific knowledge into the LLM \cite{hu2021LoRA}.

F-PEFT combines the advantages of these two fine-tuning methods. 
LoRA reconstructs the response structure of the LLM and injects relevant domain knowledge, while the Prefix module extracts information from temporal data and guides EF-LLM to provide accurate responses. 
As illustrated in Fig.\ref{How_enhance_sparse}, EF-LLM presents an illustrative diagram demonstrating the principle of how F-PEFT enhances prediction accuracy under data sparsity. 
Traditional models struggle to perform well under data sparsity. 
In contrast, LLMs, trained on large-scale corpora, inherently possess quantitative knowledge of certain sparse features and data. 
By interacting with LLMs, EF-LLM can uncover such latent knowledge. 
F-PEFT combines the strengths of both methods by utilizing numerical data for standard predictions while leveraging the textual modality to uncover similar knowledge within LLMs. 
This approach enhances the representation of sparse features and significantly improves predictive performance in scenarios with sparse data.

A similar approach of combining Prefix and LoRA was in \cite{chow2024towards} to extract time-series data and fine-tune text responses.
Despite some similarities in theory, EF-LLM and the LLM in \cite{chow2024towards} are used for different purposes and scenarios.

\subsubsection{{Implementation Procedure}}  
{Owing to its extensive knowledge reserve, the LLM inherently possesses a certain ability to understand and describe sparse features. 
However, the origin of this ability remains difficult to interpret. 
To ensure that EF-LLM can be reliably used to enhance prediction performance in sparse scenarios, we adopt a continual learning strategy when equipping it with sparse prediction capability. 
Specifically, as shown in Eq.\ref{S_imple_1}, we first perform SFT on the LLM using few-shot sparse-feature data represented purely in textual form.}

\allowdisplaybreaks
\begin{subequations}\label{S_imple_1}
\begin{align}
& w \in \mathcal{W} \ | \frac{1}{T}\sum_{t=1}^{T}\mathbb{I}(w_t=w)<\delta, \quad \delta=0.01, \label{S_imple_1a}\\
& \mathcal{T}_{s} = \left\{ t \mid w_t \in \mathcal{W}_{s} \right\}, \label{S_imple_1b}\\
& \mathcal{D}_{s} = \left\{ (\mathbf{x}_t, w_t, y_{t-1}, y_t) \mid t \in \mathcal{T}_{s} \right\}, \label{S_imple_1c}\\
& \mathcal{S}_{s} = \left\{ (q_t, a_t) \mid t \in \mathcal{T}_{s} \right\}, q_t = \phi(w_t, y_{t-1}), a_t = \psi(y_t), \label{S_imple_1d}\\
& \mathcal{L}_{SFT} = - \sum_{(q_t,a_t)\in\mathcal{S}_{s}} \sum_{i=1}^{|a_t|} \log P_{\theta} \left( a_{t,i} \mid q_t, a_{t,<i} \right), \label{S_imple_1e} \\
& \theta_s = \theta_0 + \Delta \theta_{\mathrm{SFT}}^{\mathrm{sparse}}, \label{S_imple_1f}
\end{align}
\end{subequations}
{where $\mathcal{W}$ denotes the set of all weather types in the original dataset, and $w$ represents a specific weather type selected from $\mathcal{W}$. $w_t$ denotes the weather type observed at time step $t$, while $T$ is the total number of samples in the dataset. $\mathbb{I}(\cdot)$ is the indicator function, which equals 1 if the condition inside it is satisfied and 0 otherwise. Therefore, $\frac{1}{T}\sum_{t=1}^{T}\mathbb{I}(w_t=w)$ represents the occurrence frequency of weather type $w$ in the dataset. $\delta$ is the sparsity threshold, which is set to $0.01$ in this paper. A weather type whose occurrence frequency is lower than $\delta$ is regarded as a sparse weather feature.}

{$\mathcal{W}_{s}$ denotes the sparse weather feature set selected from $\mathcal{W}$. $\mathcal{T}_{s}$ denotes the set of timestamps corresponding to sparse weather features. Specifically, if the weather type $w_t$ at time step $t$ belongs to the sparse weather feature set $\mathcal{W}_{s}$, then the timestamp $t$ is included in $\mathcal{T}_{s}$.}

{$\mathcal{D}_{s}$ denotes the constructed sparse dataset. For each timestamp $t \in \mathcal{T}_{s}$, $\mathbf{x}_t$ represents the numerical time-series features, $w_t$ represents the corresponding sparse weather type, $y_{t-1}$ denotes the ground-truth output at the previous time step, and $y_t$ denotes the ground-truth output at the current time step. In this stage, $y_{t-1}$ is used as the persistence forecast reference value for constructing the sparse-feature-oriented textual instruction.}

{$\mathcal{S}_{s}$ denotes the sparse instruction tuning dataset constructed from $\mathcal{D}_{s}$. Each sample in $\mathcal{S}_{s}$ consists of an input prompt $q_t$ and a ground-truth answer $a_t$. The function $\phi(\cdot)$ denotes the prompt construction function, which converts the sparse weather type $w_t$ and the previous-step reference value $y_{t-1}$ into a textual input prompt $q_t$. The function $\psi(\cdot)$ denotes the answer construction function, which converts the current ground-truth output $y_t$ into the target textual response $a_t$. An example of $\mathcal{S}_{s}$ is as follows:}
\textcolor{blue}{\itshape input prompt: ``I asked you about the power range and power forecast of PV. your answer was: power will be 11$\sim$12\% of the rated energy, power will be 97.5. Based on your previous answer, now the weather is: `clear turning to heavy rain'. How will this affect the results, can you give me a new power range forecast.'' ground truth: ``power will be 7$\sim$8\% of the rated energy.''}

{$\mathcal{L}_{SFT}$ denotes the SFT loss for the sparse instruction tuning stage. $P_{\theta}(\cdot)$ represents the token generation probability parameterized by the LLM parameters $\theta$. $a_{t,i}$ denotes the $i$-th token in the ground-truth answer $a_t$, and $a_{t,<i}$ denotes all tokens before $a_{t,i}$. $|a_t|$ represents the total number of tokens in $a_t$. The loss function minimizes the negative log-likelihood of generating the ground-truth answer $a_t$ given the input prompt $q_t$.
$\theta_0$ denotes the parameters of the original pre-trained LLM, while $\theta_s$ denotes the parameters of the LLM after sparse-feature-oriented SFT. 
$\Delta \theta_{\mathrm{SFT}}^{\mathrm{sparse}}$ represents the parameter update learned from the sparse SFT stage. 
Through this process, the original LLM is adapted into a sparse-aware LLM, which is then used as the initialized LLM base for the subsequent multimodal fine-tuning stage.}

{After completing the operations in Eq.~\ref{S_imple_1}, EF-LLM further performs continual knowledge injection into the LLM through the F-PEFT method, following the continual learning scheme described in Eq.~\ref{Su_2} and Fig.~\ref{continual_learning_LoRA}. 
This process enables the model to generalize to the complete dataset, thereby achieving the objective of using the LLM to provide sparse-information complementarity for time-series information. 
Therefore, this procedure constitutes the complete implementation process of the principle in Fig.~\ref{How_enhance_sparse}.
Such sparse weather features are inherently represented in textual form, with diverse expression styles and semantic content, making them difficult to be effectively described by conventional numerical prediction models. 
Therefore, using an LLM to enhance such information has unique advantages and provides a reliable complementary approach to time-series information.}

\subsection{{Hallucination Detection}} \label{SEC_hallucination}
{Since the output of EF-LLM is textual, EF-LLM uses the semantic cosine similarity between the generated response and the target response to identify and quantify parts that do not follow the expected fine-tuned response format or deviate substantially from the target semantic content. 
A low semantic cosine similarity indicates textual hallucination \cite{mikolov2013efficient, reimers2019sentence}.}

{Semantic similarity can measure the degree of closeness in meaning between multiple sentences. 
By calculating the cosine of the angle between these high-dimensional vectors, their similarity can be assessed, as in Eq.\ref{S_h}.}
\begin{equation}\label{S_h}
\wp(E^{LLM}, O^{LLM}) = 1 - \frac{E^{LLM} \cdot O^{LLM}}{||E^{LLM}|| \cdot ||O^{LLM}||},
\end{equation}
{where $E^{LLM}$ is the high-dimensional vector representation of the expected EF-LLM output.
$||E^{LLM}||$ and $||O^{LLM}||$ are the norm of the corresponding vectors.
$\wp$ is cosine similarity.
The larger the value of $\wp$, the greater the similarity of the sentences.}

\subsection{{Numerical Output Reliability Analysis}} \label{SEC_reliability}
{Since EF-LLM’s textual outputs also include certain numerical information, such as predicted proportion intervals (e.g., 20--21$\%$) and numerical values (e.g., 24.4), the stability of these outputs needs to be analyzed to ensure the reliability of the generated textual responses. 
Therefore, we conducted a numerical output reliability analysis based on ANOVA, a statistical analysis method.
Specifically, 100 EF-LLM responses are collected for the same input, and the numerical outputs are extracted from the text responses. 
ANOVA is then applied to these extracted numerical results to quantify their fluctuation, as shown in Eq.~\ref{S_2} \cite{st1989analysis}.
$\chi_{i}$ is the i-th Observations.
$\bar{\chi}$ is the mean value.
$n$ is observation number.
$k$ is experimental set number.
$n_j$ is observation number in the j-th experimental sets.
SSB, SSW are the sum of squares between and within groups respectively, 
$F$ is the F-statistic value.}
\allowdisplaybreaks
\begin{subequations}\label{S_2}
\begin{align}
&\text{SSB} = \sum^k_{j=1} n_j(\chi_j - \bar{\chi})^2, \label{S_2a}\\
&\text{SSW} = \sum^k_{j=1} \sum^{n_j}_{i=1} (\chi_{ij} - \bar{\chi_j})^2, \label{S_2b}\\
&\text{MSB} = \frac{\text{SSB}}{k-1}, \label{S_2c}\\
&\text{MSW} = \frac{\text{SSW}}{n-k}, \label{S_2d}\\
&F = \frac{\text{MSB}}{\text{MSW}}, \label{S_2e}
\end{align}
\end{subequations}

\section{Case Study}    \label{SEC_CaseStudy}
In the EF-LLM base, EF-LLM has integrated knowledge from three energy forecasting scenarios: load , PV, and wind power forecasting. 
Additionally, for each scenario, EF-LLM has included comprehensive automated operation knowledge, including pre-forecast data preparation guidance, feature engineering, prompt engineering, multimodal time-series forecasting, and post-forecast decision support.
The entire forecasting task is completed through interactions with EF-LLM.

Case study is divided into four modules:
\ref{CaseStudy_1}: Data introduction.
\ref{CaseStudy_2}: AI-Assisted automation with EF-LLM: demonstrating how EF-LLM achieves AI-assisted automation, including comparisons of EF-LLM responses before and after F-PEFT and the impact of continual knowledge injection on EF-LLM's response quality.
\ref{CaseStudy_3}: EF-LLM performance under feature or data sparsity.
\ref{CaseStudy_4}: Hallucination detection \& quantitative analysis: exploring how the hallucination rate in forecasting tasks varies with changes in the multi-task coefficient $\varpi$.

GLM4-9b is used as the pretrained LLM base \cite{GLM4}, finetuned with an L20 48GB GPU, gradient: 1e-3, 1-3 epochs.

\subsection{Data \& Scenarios} \label{CaseStudy_1}
The load data was provided by Ausgrid, consisting of data from 300 home customers in Australian NSW state.
The PV data was from China Southern Grid, a 798 kWh PV plant located in Ningbo, China, at coordinates longitude 121.2944, latitude 29.32238, in 2022. 
The wind data is from Kaggle \cite{jorge_2021}.

\begin{figure} [!htb]
        \subfigure[Load Data Box Plot.] { 
            \label{load_BoxPlot}     
            \includegraphics[width=3.5in]{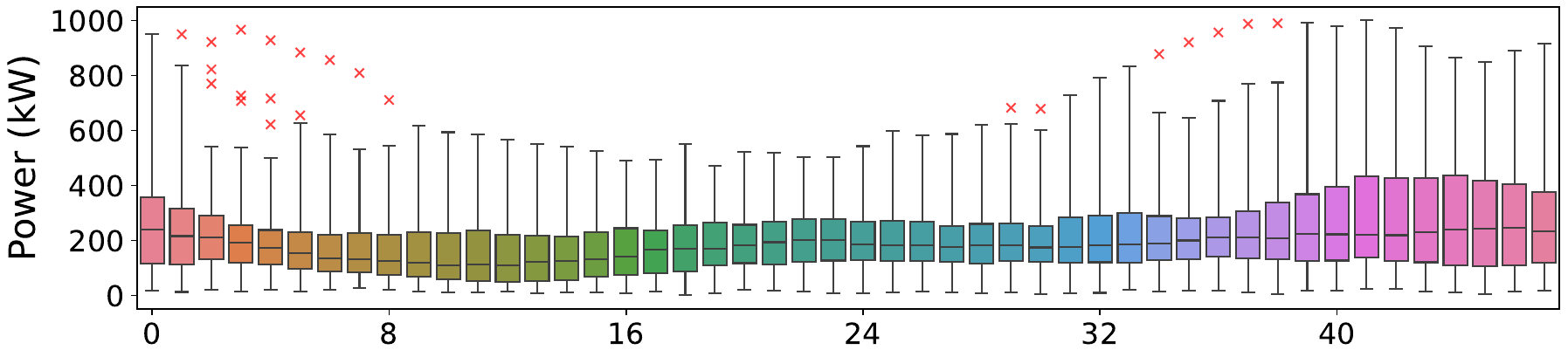} 
        } 
        \subfigure[PV Power Data Box Plot.] { 
            \label{PV_BoxPlot}     
            \includegraphics[width=3.5in]{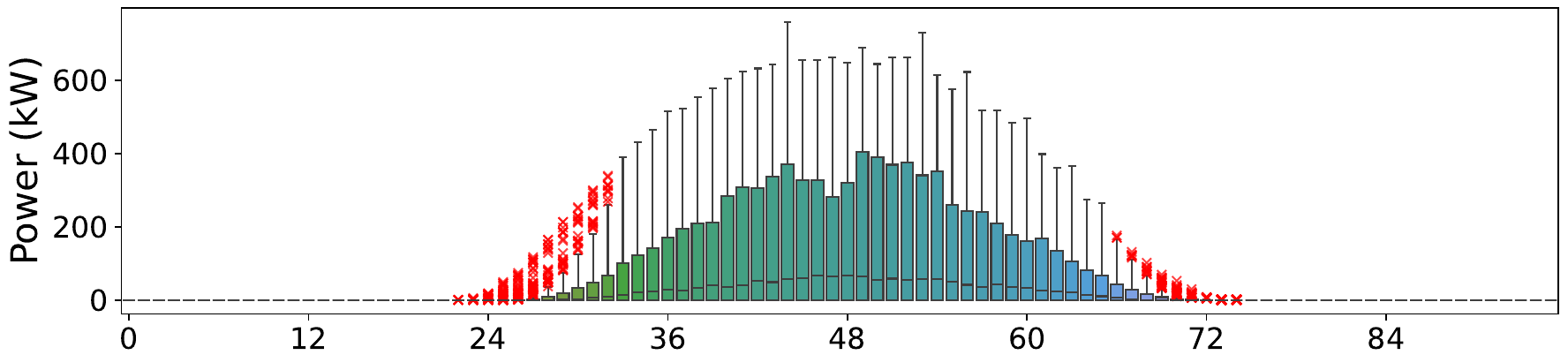}  
        }       
        \subfigure[Wind Power Data Box Plot.] { 
            \label{wind_BoxPlot}     
            \includegraphics[width=3.5in]{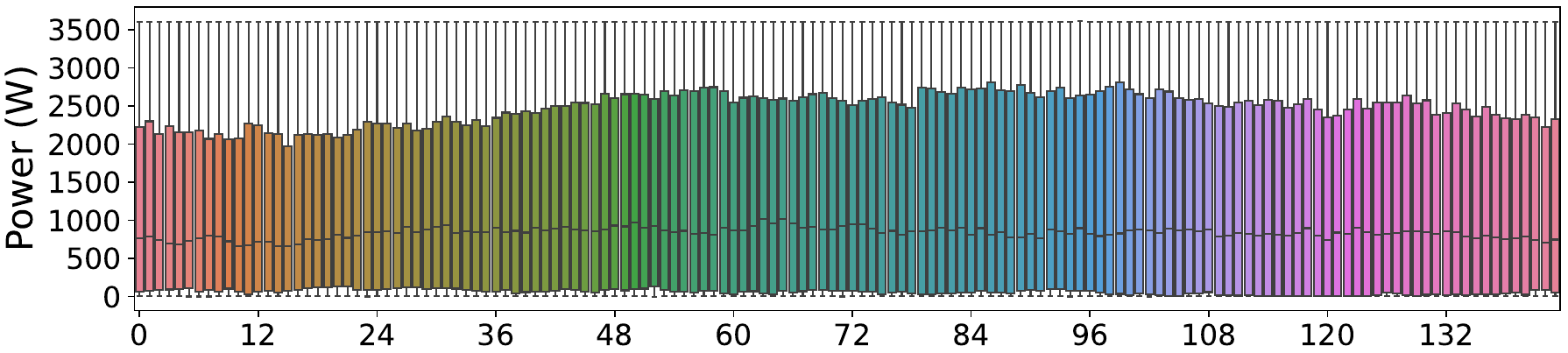} 
        } 
        \caption{Box Plots of Chosen Scenario Data in Energy Systems.}     
        \label{box_plots}     
    \end{figure}

Fig.\ref{box_plots} shows the box plot distribution of the data for the three selected scenarios over the course of a day.

\subsection{AI-assisted Automation} \label{CaseStudy_2}
EF-LLM enables users to achieve AI-assisted automation, with a case study presented in the Appendix. The appendix of the paper can be accessed via a subsequent link \cite{DropBox}. 
Users do not need years of experience or specialized expertise; by interacting with EF-LLM step by step, no manual intervention is required. 
EF-LLM can independently complete the corresponding tasks.

\begin{figure} [!htb]
	\centering
	\includegraphics[width=5.5in]{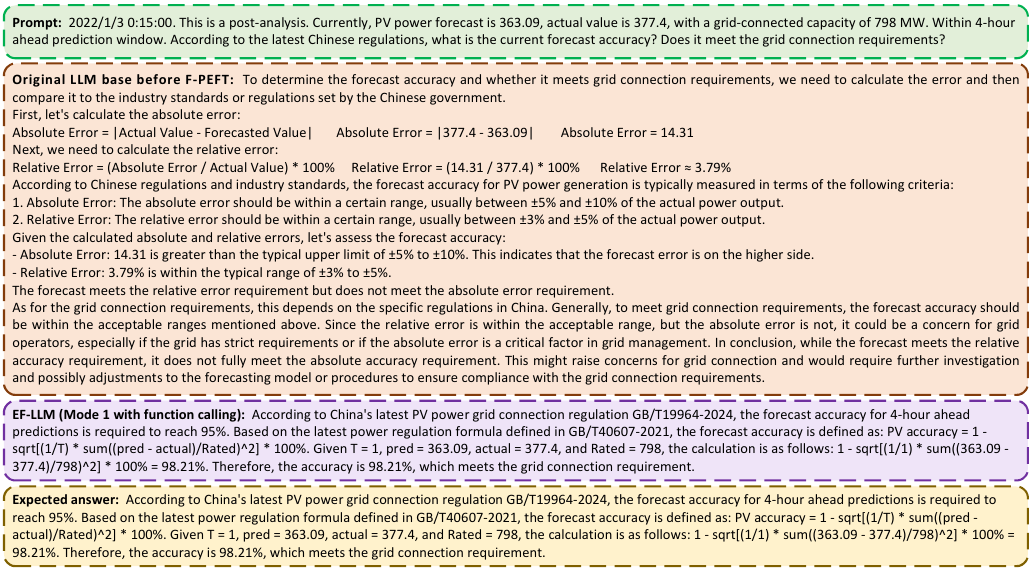}
	\caption{Comparison: decision support task before and after F-PEFT.}
	\label{compare_post_analysis}
\end{figure}

Before F-PEFT, the pre-trained LLM base cannot handle domain-specific tasks. 
Fig.\ref{compare_post_analysis} shows the results of a decision support task. 
Before F-PEFT knowledge injection, EF-LLM was unable to handle decision support tasks effectively. 
The responses not only misunderstood the problem but also produced incorrect calculations. 
Due to the pre-trained LLM base's inability to address the given problems, its responses exhibited low semantic similarity to the expected answers. 
As shown in Fig.\ref{compare_post_analysis}, the cosine similarity between the original pre-trained LLM base's response and the expected response was only 0.389, indicating significant semantic deviation.

After F-PEFT, the semantic similarity between EF-LLM's responses and the expected responses often reached 100\%, fully aligning with the expected answers. 
This improvement is attributed to the techniques employed during the fine-tuning of EF-LLM for this task, including fixed response formats and function calling. 
For example, in scenarios requiring highly precise mathematical calculation outputs, such as 98.21\% shown in Fig.\ref{compare_post_analysis}, function calling was essential to eliminate any errors.

The four tasks injected into EF-LLM utilized the techniques shown in Table \ref{TB_AA_1}.
Data preparation: the pre-trained LLM base already had relevant knowledge. 
To preserve its original knowledge content, EF-LLM did not constrain the response format during F-PEFT \cite{tam2024let}. 
Instead, EF-LLM is fine-tuned to include specific knowledge about the data requirements for selected scenarios.
Feature \& prompt Engineering: the pre-trained LLM base had sufficient knowledge reserves, so no additional knowledge was injected. 
Instead, function calling mechanisms were introduced, enabling EF-LLM to select appropriate feature processing algorithms based on the prediction scenario.
Prediction tasks: for the selected scenarios, EF-LLM performed multimodal time-series forecasting independently without using external functions.
Decision support: as demonstrated in Fig.\ref{compare_post_analysis}, to achieve fully expected responses for mathematical problems, all three techniques were employed: restrict format, function calling, and knowledge injection.

  \begin{table} [!htb]
    \centering
    \caption{Techniques Used For Different Tasks}
        \label{TB_AA_1}
    \renewcommand\arraystretch{1.5}
    \begin{tabular} {c|c|c|c}
        \hline
        \hline
         Techniques   &  Restrict format  &  Function call & Inject knowledge  \\ 
        \hline
        prepare data  &   $\times$  &   $\times$  &  $\surd$ \\ 
        \hline
        F\&P engineering  &   $\times$   &   $\surd$  &  $\times$\\
        \hline
        prediction  &   $\surd$   &   $\times$  &  $\surd$\\
        \hline
        decision support  &   $\surd$   &   $\surd$  &  $\surd$\\
        \hline
        \hline
    \end{tabular}  
  \end{table}

The semantic similarity between the pre-trained LLM base and EF-LLM's responses to the target answers improved across various tasks after knowledge injection, as in Table \ref{TB_AA_2}.
In the data preparation task, since the pre-trained LLM base already possessed relevant knowledge, it was able to achieve relatively high semantic accuracy in its responses even before F-PEFT. 
However, knowledge injection further significantly enhanced semantic accuracy, enabling EF-LLM to provide more precise responses tailored to specific scenarios.
For highly specialized tasks such as time-series forecasting and decision support, the pre-trained LLM base lacked the necessary domain knowledge, resulting in substantial discrepancies between its responses and the expected answers. 
After knowledge injection, semantic accuracy improved significantly. 
An average semantic similarity above 0.9 with the target responses indicates that the overall format of the sentences is completely correct, though numerical deviations still exist. 
This is because EF-LLM, as a predictive model, cannot produce prediction results identical to actual values.
For purely mathematical problems, EF-LLM's responses are nearly identical to the expected answers, with only minor deviations in a few words.

  \begin{table} [!htb]
    \centering
    \caption{Average Cosine Similarity Comparison to Targets Before And After F-PEFT}
        \label{TB_AA_2}
    \renewcommand\arraystretch{1.5}
    \begin{tabular} {c|c|c}
        \hline
        \hline
               &  Pre-trained LLM  &  EF-LLM  \\   
        \hline
        data preparation  &  0.781 &  0.844  \\ 
        \hline
        prediction  &  0.353  &  0.947   \\
        \hline
        decision support  &  0.367  &  0.999   \\
        \hline
        \hline
    \end{tabular}  
  \end{table}

New knowledge can be continuously injected into EF-LLM with LoRA. 
Table \ref{TB_AA_3} demonstrates the effects of continual knowledge injection. 
All the data is divided into two parts, injecting the first half initially and then supplementing the remaining half through LoRA-based continual learning.
For tasks with non-fixed formats, such as data preparation, continual knowledge injection significantly improved semantic accuracy. 
However, the improvement was not as pronounced as injecting all the knowledge at once. 
This suggests that subsequent knowledge injection may have a slight adverse impact on previously injected knowledge.
For fixed-format prediction tasks, while continual learning led to some improvement in semantic similarity, the effect was relatively minor.
For decision support, as long as the injected knowledge enables EF-LLM to recognize users' prompts and trigger corresponding functions, no additional knowledge injection is necessary. 

  \begin{table} [!htb]
    \centering
    \caption{Average Similarity to Standard Answers Before And After Continual Learning}
        \label{TB_AA_3}
    \renewcommand\arraystretch{1.5}
    \begin{tabular} {c|c|c|c}
        \hline
        \hline
               &  EF-LLM base  &  updated EF-LLM  &  EF-LLM(direct) \\   
        \hline
        data preparation  &  0.813  &  0.839  &  0.844  \\ 
        \hline
        prediction  &  0.945  &  0.947  &  0.947 \\
        \hline
        decision support  &  0.999  &  0.999  &  0.999 \\
        \hline
        \hline
    \end{tabular}  
  \end{table}

\subsection{Forecasting Performance Under Data Sparsity} \label{CaseStudy_3}
{This subsection evaluates the prediction performance of EF-LLM under data sparsity from three perspectives. The first test investigates the prediction performance with sparse feature categories, as shown in Table \ref{TB_1}. This experiment mainly examines whether EF-LLM can benefit from the intrinsic characteristics of LLMs, including strong representation capability, fast convergence with fewer training steps, and reduced risk of overfitting when only limited sparse data are available.
The second test compares the prediction performance before and after text-enhanced sparse features, which corresponds to the continual-learning-based sparse feature enhancement method introduced in Section \ref{SEC_sparse}, as shown in Table \ref{TB_2}. In this method, sparse data are first injected into the LLM in a step-by-step manner and then generalized to the complete dataset. Therefore, this test is designed to evaluate the prediction capability of the continually enhanced EF-LLM under CoT-based text-enhanced sparse features. In particular, it examines whether the model can leverage text-enhanced information to compensate for the information loss caused by sparse features.
The third test evaluates the transfer learning performance under data sparsity. This experiment focuses on the generalization capability of EF-LLM when the source data are limited or the target scenario contains insufficient data, as shown in Table \ref{TB_3}. By conducting this test, we further examine whether EF-LLM can maintain strong cross-task or cross-scenario adaptability under sparse-data conditions.}

In the accuracy experiment, EF-LLM's performance was evaluated under feature category sparsity.
For load prediction, the features considered included holidays, temperature, dew point, and wind speed. 
For PV, wind speed, sunlight, and weather type were considered. 
For wind, wind speed and direction were considered.
Feature category sparsity is a common type of data sparsity.
Table \ref{TB_1} compares EF-LLM's performance under feature sparsity with other neural networks, including classical LSTM and state-of-the-art models as iTF (iTransformer) and TN (TimesNet). 
EF-CLS represents EF-LLM classification task, and EF-REG represents EF-LLM regression task.
{To further strengthen the evaluation against recent LLM-based forecasting methods in the energy domain, we additionally include a STELLM-style baseline for comparison. 
Since the official implementation of STELLM is not publicly available, we construct a reproducible variant by adapting the open-source Time-LLM framework \cite{jin2023time} following the methodological descriptions reported in STELLM \cite{wu2024stellm}, including patch-level GPT-based representation learning, seasonal-trend decomposition, and linear numerical forecasting heads.}

\begin{table} [!htb] 
  	\centering
  	\caption{Prediction Performance Under Feature Sparsity}
        \label{TB_1}
	\renewcommand\arraystretch{1.5}
  	\begin{tabular} {c|c|c|c|c|c|c}
  		\hline
  		\hline
  		Load  &  EF-CLS   &  EF-REG  &  LSTM  &  iTF  &  TN  &  {STELLM-style}  \\   
  		\hline
  		MAE (kW) &   \textcolor{red}{18.14}    &   24.34   &   24.28   &    20.34   &    \textcolor{blue}{19.32}   &   {22.73} \\ 
  		\hline
  		RMSE (kW) &   \textcolor{red}{20.63}    &   33.13   &   32.56   &   24.62   &   \textcolor{blue}{23.41}   &   {27.19} \\
  		\hline
  		\hline
  		PV &  EF-CLS   &  EF-REG  &  LSTM  &  iTF  & TN   &   {STELLM-style}  \\   
  		\hline
  		MAE (kW)  &  \textcolor{red}{15.07}   &  16.81  &  17.71  &   16.69  &  \textcolor{blue}{16.16}   &  {18.72}\\ 
  		\hline
  		RMSE (kW) &  \textcolor{red}{38.18}   &  46.79  &  \textcolor{blue}{41.44}  &  43.99  &  45.14   &   {56.48}  \\
  		\hline
  		\hline
  		Wind  &  \makecell{EF-CLS}   &  \makecell{EF-REG}  &  LSTM  &  iTF  & TN   &   {STELLM-style}  \\   
  		\hline
  		MAE (W)  &  100.34  &   132.22   &   128.28   &   \textcolor{blue}{89.68}   &    \textcolor{red}{88.57}   &   {103.13}  \\ 
  		\hline
  		RMSE (W)  &  193.46  &   254.23   &    246.05    &   \textcolor{blue}{174.77}   &     \textcolor{red}{153.20}   &   {213.51} \\ 
            \hline
            \hline
  	\end{tabular}  
  \end{table}

The accuracy of EF-LLM classification is measured by comparing the median of the prediction interval with the ground truth. Considering the uncertainty in LLM responses, the prediction accuracy of EF-LLM is obtained by averaging 50 response samples. It is evident that under feature sparsity, EF-LLM demonstrates excellent prediction performance, surpassing various benchmarks in load and PV scenarios. However, the time-series forecasting framework based on LLMs is influenced by the weight of original textual knowledge and the limit on training epochs, which can constrain accuracy in some forecasting tasks. For example, in wind prediction, its accuracy is inferior to iTransformer and TimesNet, but it still significantly outperforms LSTM.
{Because the training stage lacks reinforcement of sparse features, the performance of the STELLM-style approach under data sparsity is clearly weaker than that of EF-CLS. The limitation of conventional time-series neural networks in insufficiently representing sparse features still persists in STELLM.}

To further examine the relationship between the two prediction targets, the internal consistency was analyzed between the classification-style interval prediction and the numerical value prediction. The results show a clear dependency between these two outputs. When the predicted interval matches the ground-truth interval, the predicted numerical value remains within the predicted interval in 97.87$\%$ of cases. 
In contrast, when the predicted interval does not match the ground-truth interval, this proportion drops to 40.32$\%$. 
This large gap indicates that the interval prediction serves as an important structural anchor for the numerical output. 
In other words, EF-LLM exhibits much stronger internal consistency when the classification-style interval prediction is correct, which provides additional support for the observation that the model is relatively more robust in interval-based classification-style prediction than in exact numerical regression, especially under sparse conditions.

Although EF-LLM does not achieve the best performance across all forecasting tasks, it offers two key advantages. First, EF-LLM integrates time-series prediction across multiple scenarios, including load, PV, and wind, allowing it to handle multiple forecasting tasks simultaneously. Second, EF-LLM possesses function-calling capabilities. If it cannot provide the best prediction results in certain scenarios, it can still call models such as TimesNet or iTransformer to generate predictions. Therefore, EF-LLM not only leverages its own strengths, but also integrates the advantages of various forecasting models.

\begin{table} [!htb]
  	\centering
  	\caption{Prediction Performance Under CoT-enhanced Sparse Feature}
        \label{TB_2}
	\renewcommand\arraystretch{1.5}
  	\begin{tabular} {c|c|c|c|c|c|c}
  		\hline
  		\hline
  		load  &  \makecell{1-C}   &  \makecell{1-R}  &  \makecell{2-C}  &  \makecell{2-R}  &  \makecell{EF-C}  &  \makecell{EF-R} \\   
  		\hline
  		MAE (kW)  &  18.65  &  24.45  &  \textcolor{red}{17.93}  &  24.27  &  \textcolor{blue}{18.14}  &  24.34 \\ 
  		\hline
  		RMSE (kW) &  21.47  &  33.23  &  \textcolor{red}{19.92}  &  32.69  &  \textcolor{blue}{20.63}  &  33.13  \\ 
  		\hline
  		\hline
  		PV  &  \makecell{1-C}   &  \makecell{1-R}  &  \makecell{2-C}  &  \makecell{2-R}  &  \makecell{EF-C}  &  \makecell{EF-R} \\   
  		\hline
  		MAE (kW)  &  16.52  &  17.62  &  \textcolor{red}{14.62}  &  17.45  &  \textcolor{blue}{15.07}  &  16.81 \\ 
  		\hline
  		RMSE (kW) &  42.36  &  47.41  &  \textcolor{red}{37.93}  &  45.33  &  \textcolor{blue}{38.18}  &  46.79 \\ 
            \hline
            \hline
  		wind  &  \makecell{1-C}   &  \makecell{1-R}  &  \makecell{2-C}  &  \makecell{2-R}  &  \makecell{EF-C}  &  \makecell{EF-R} \\   
  		\hline
  		MAE (W)  &  108.88  &  138.05  &  \textcolor{red}{96.95}   & 124.23   &  \textcolor{blue}{100.34}  &  132.22 \\ 
  		\hline
  		RMSE (W) &  213.93  &  266.91  &  \textcolor{red}{188.02}  & 238.30   &  \textcolor{blue}{193.46}  &  254.23 \\ 
            \hline
  		\hline
  	\end{tabular}  
  \end{table}

To further demonstrate EF-LLM's performance under sparse features, sparse feature enhancement reasoning based on the Chain of Thought (CoT) framework was conducted to test EF-LLM's capabilities. 
Some features are very sparse throughout the dataset. 
For example, in PV data, features like "heavy rain turning to clear," "clear turning to heavy rain," "haze," and "cloudy turning to overcast" are rare. 
In 2022, "heavy rain turning to clear" only occurred three times, "clear turning to heavy rain" occurred four times. 
Such features are challenging to represent in traditional forecasting models. 
However, the pretrained knowledge of LLMs from extensive corpora enables them to better quantify such sparse features as "heavy rain turning to clear".

\begin{figure} [!htb] 
	\centering
	\includegraphics[width=5.5in]{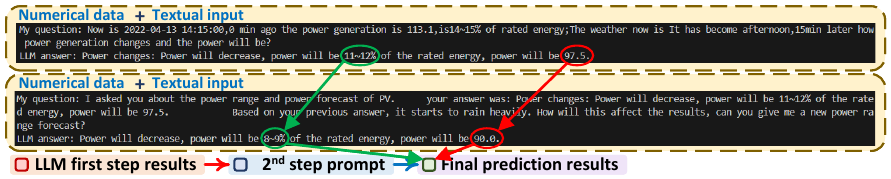}
	\caption{EF-LLM CoT prompting inference process.}
	\label{two_steps}
  \end{figure}

To enhance such features, CoT reasoning was applied as shown in Fig.\ref{two_steps}. 
In the first step, only time-series data was provided, weather features were not mentioned in the prompt.
Predictions were based solely on time-series data and the text used to trigger corresponding prediction tasks.
Second step, CoT reasoning was used to supplement weather features, enhancing sparse feature quantification and generating refined prediction results.
Table \ref{TB_2} shows the prediction results, where 1-C and 1-R mean classification and regression tasks of EF-LLM in the first step, respectively.
Red text indicates the best result, while blue represents the second-best.
2-C and 2-R mean classification and regression tasks in the second step with CoT enhancement. 
EF-C and EF-R denote the performance of EF-LLM when all information is directly provided in a single step for classification and regression reasoning.

  \begin{figure} [!htb]
	\centering
	\includegraphics[width=5.5in]{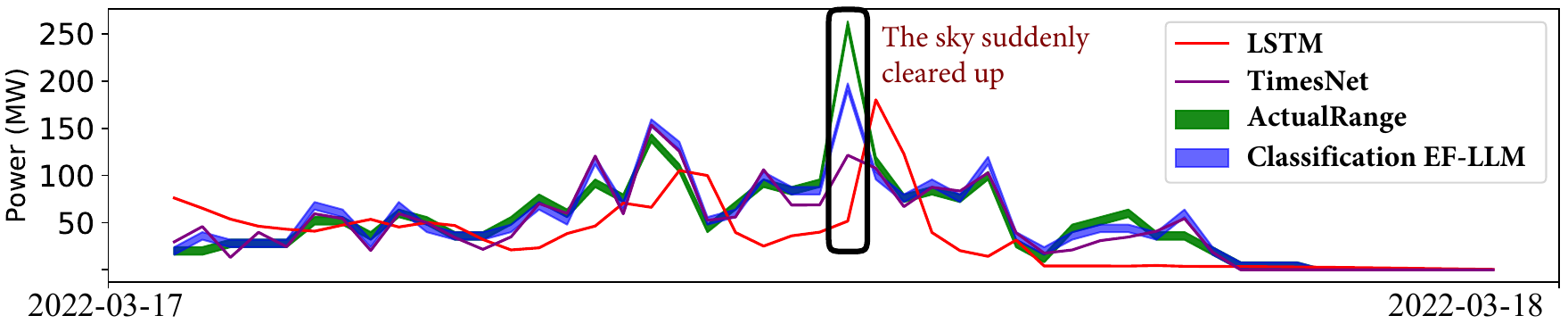}
	\caption{EF-LLM Enhanced PV Power Prediction Under Feature Sparsity.}
	\label{Extreme_compare}
  \end{figure}

After CoT reasoning enhances sparse features, the prediction accuracy surpasses the results obtained from direct one-step predictions. 
This demonstrates that CoT can improve the importance of the supplemented sparse features in the prompt, thereby achieving an enhancement effect. 
As shown in Fig.\ref{Extreme_compare}, a case for enhancing "heavy rain turning to clear" illustrates this point. 
Even TimesNet struggles to capture such features, whereas EF-LLM exhibits exceptional performance under sparse feature conditions.

\begin{table} [!htb]
  	\centering
  	\caption{Transfer Learning Performance Under Data Sparsity}
        \label{TB_3}
	\renewcommand\arraystretch{1.5}
  	\begin{tabular} {c|c|c|c|c|c|c}
  		\hline
  		\hline
  		load  &  EF-CLS   &  \makecell{EF-REG}  &  \makecell{LSTM}  &  \makecell{iTF}  &  \makecell{TN} &  {STELLM-style}  \\   
  		\hline
  		MAE (kW)  &  \textcolor{red}{25.17}  &  32.72  &  31.81  &  \textcolor{blue}{27.98}  &  29.08  &  {31.62}  \\ 
  		\hline
  		RMSE (kW) &  \textcolor{red}{33.83}  &  53.88  &  49.71  &  40.28  &  \textcolor{blue}{37.92}  &  {51.56}    \\ 
  		\hline
  		\hline
  		PV  &  EF-CLS   &  EF-REG  &  \makecell{LSTM}  &  \makecell{iTF}  &  \makecell{TN}  &  {STELLM-style}\\   
  		\hline
  		MAE (kW)  &  \textcolor{red}{10.32}  &  16.73  &  \textcolor{blue}{13.83}  &  19.93  &  24.40  &   {15.13}  \\ 
  		\hline
  		RMSE (kW) &  \textcolor{blue}{20.40}   &  24.39  &  \textcolor{red}{18.27}   &  33.34  &  40.19 &   {22.23} \\ 
            \hline
            \hline
            wind  &  \makecell{EF-CLS}   &  \makecell{EF-REG}  &  \makecell{LSTM}  &  \makecell{iTF}  &  \makecell{TN}  &  {STELLM-style} \\   
  		\hline
  		MAE (W)  &  \textcolor{blue}{116.71}  &  145.33  &  \textcolor{red}{134.17}  &  168.26  &  138.12   &   {154.53} \\ 
  		\hline
  		RMSE (W) &  \textcolor{blue}{228.55}  &  272.45  &  \textcolor{red}{252.86}  &  310.21  &  254.58   &   {295.10} \\ 
            \hline
            \hline
  	\end{tabular}  
  \end{table}

Sparse training data issue is also a common challenge. 
For example, transfer learning was conducted across three scenarios:
(1) Load: From February to April 2020, during the onset of COVID-19, Australia's load characteristics began to change. 
At this time, there was no similar training dataset available, and transfer learning under sparse data conditions was an option. 
EF-LLM, which had been injected with load forecasting knowledge from 2018, was utilized and transferred to this period for continual knowledge injection.
(2) PV: A 298kWh PV plant near another 798kWh PV plant lacked sufficient data. 
EF-LLM, trained on data from the 798kWh PV plant, was transferred to the 298kWh PV plant for continual learning.
(3) Wind: The wind data was divided into two parts. 
EF-LLM was trained on the first half and then transferred to the second half for continual transfer learning.

The prediction results are shown in Table \ref{TB_3}. 
EF-LLM demonstrated outstanding continual transfer learning capabilities under sparse data conditions, {significantly outperforming iTransformer, TimesNet, and STELLM-style.}  
This highlights EF-LLM's transfer learning performance under data sparsity.

\subsection{Multitask Learning Based Hallucination Detection}\label{CaseStudy_4}
\begin{figure} [!htb]
	\centering
	\includegraphics[width=5.5in]{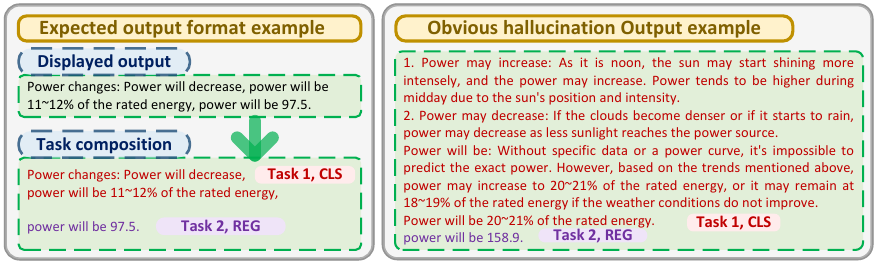}
	\caption{Typical Hallucination Example.}
	\label{coef_influence}
  \end{figure}

{For hallucination detection, this work employs multi-task learning and semantic cosine similarity calculations to quantitatively reflect the proportion of hallucinations in text responses.}
As in Eq.\ref{S_h}, when the semantic difference $\wp$ between EF-LLM’s response and the standard response falls below a certain threshold, the output is considered as a hallucination. 
The value of $\wp$ that is deemed indicative of hallucination depends on the user’s tolerance for semantic deviations in the current task.
Taking PV prediction as an example, $\wp$ below 0.8 indicates that EF-LLM has not followed the expected format, resulting in hallucination.
To make hallucination more visually apparent, a multi-task learning approach is applied. 
By adjusting the multi-task coefficient $\varpi$, hallucinated response proportion within all EF-LLM responses changes, revealing a clear pattern, as shown in Fig.\ref{coef_influence} and Table \ref{TB_hallu}.
MAE-C is MAE of CLS task, MAE-R is MAE of REG task, HP-C is hallucination proportion of CLS task, HP-R is hallucination proportion of REG task.
The larger $\varpi$, the higher the CLS task weight and the lower the REG task weight.
This shows that, regardless of the task, when its task weight decreases, hallucinations start to appear. 
\begin{table} [!htb]
    \centering
    \caption{$\varpi$ Influence On Accuracy \& Hallucination Proportion}
        \label{TB_hallu}
    \renewcommand\arraystretch{1.5}
    \begin{tabular} {c|c|c|c|c|c|c|c}
        \hline
        \hline
        load $\wp$  &  0.2  &  0.3  &  0.4  &  0.5  &  0.6  &  0.7  &  0.8 \\   
        \hline
        MAE-C  &  19.9  &  17.4  &  16.8  &  15.1  &  15.0  &  15.1  &  15 \\ 
        \hline
        MAE-R  &  16.8  &  16.6  &  16.7  &  16.8  &  17.2  &  17.4  &  17.8 \\
        \hline
        HP-C/\%  &   22.8  &  7.5   &  2.1    &  0  &  0  &  0  &  0 \\ 
        \hline
        HP-R/\%  &  0  &  0  &  0  &   0 &  0  &  1.3  &  14.9 \\ 
        \hline
        \hline
        PV $\wp$ &  0.2  &  0.3  &  0.4  &  0.5  &  0.6  &  0.7  &  0.8 \\   
        \hline
        MAE-C  &  21.1  &  20.6  &  19.2  &  18.1  &  18.3  &  18.1 &  18.1 \\ 
        \hline
        MAE-R  &  23.9  &  24.4  &  24.2  &  24.3  &  24.7  &  24.3  &  24.54\\
        \hline
        HP-C/\%  &   24.4  &  8.2   &  2.5    &  0  &  0  &  0  &  0\\ 
        \hline
        HP-R/\%  &  0  &  0  &  0  &  0  &  0  &  0.9  &  6.1\\  
        \hline
        \hline
        wind $\wp$ &  0.2  &  0.3  &  0.4  &  0.5  &  0.6  &  0.7  &  0.8 \\   
        \hline
        MAE-C  &  128.3  &  118.0  &  106.5  &  100.3  &  99.8  &  97.2  &  96.3 \\ 
        \hline
        MAE-R  &  129.6  &  131.5  &  130.2  &  132.2  &  133.9  &  136.3  &  138.1\\
        \hline
        HP-C/\%  &   30.7   &   21.2   &   8.5  &  0   &   0  &   0  &  0  \\ 
        \hline
        HP-R/\%  &  0  &  0  &  0  &  0  &  2.6  &  12.5  &  24.8   \\  
        \hline
        \hline
    \end{tabular}  
  \end{table}

Fig.\ref{coef_influence} shows an example of hallucinations that occurred in the experiments. 
It can be seen that such hallucinations typically manifest in the form of self-reasoning.
This indicates that when hallucination occurs, the prediction results of EF-LLM no longer depend on multimodal input but rely solely on the textual description. 
Due to the model's insufficient ability to process multimodal data during hallucination, the prediction accuracy also decreases. 
The lower the task weight, the higher the hallucination proportion, and the higher prediction MAE.

From this experimental result, it can be concluded that when fine-tuning EF-LLM for prediction tasks, no unexpected, non-standard output will occur under normal circumstances. 
However, if EF-LLM cannot extract enough information from the input data, it responds with unexpected, non-standard output, and the $\wp$ will drop under the threshold. 
When this happens, a certain portion of the data has become ineffective, and the current prediction results is a hallucination.

\subsection{{ANOVA For Numerical Output Reliability Analysis}}
LLM-based responses are generated through probability-based token selection; therefore, the numerical forecasting component in EF-LLM outputs may exhibit some degree of cross-run variability even under identical inputs. 
To evaluate the stability of this numerical output component, ANOVA was conducted after EF-LLM inference. 
Specifically, for each scenario, identical inputs were used and performed 100 repeated inference runs with EF-LLM, as summarized in Table~\ref{TB_ANOVA}.
The numerical intervals and values were then extracted from the generated text responses and analyzed using ANOVA.

  \begin{table} [!htb]
    \centering
    \caption{ANOVA Results For Energy Forecasting Scenarios}
        \label{TB_ANOVA}
    \renewcommand\arraystretch{1.5}
    \begin{tabular} {c|c|c|c}
        \hline
        \hline
        ANOVA  &  Load  &  PV  &  wind \\   
        \hline
        F-statistic  &  0.00265 &  0.00473  &  0.00154 \\ 
        \hline
        P-value  &  0.9999  &  0.9999   &  0.9999  \\
        \hline
        \hline
    \end{tabular}  
  \end{table}

The ANOVA of EF-LLM across different scenarios shows that the F-statistic is very small. 
This indicates that there is no significant difference in the 100 inference results of EF-LLM given the same multimodal input data. 
The outputs of EF-LLM are statistically stable for identical inputs; despite some uncertainty, the variations are minimal.
Additionally, the P-value is close to 1, suggesting the group means across the 100 inference tests are nearly equal for the entire dataset. 
As long as the input data remains constant, the prediction results exhibit little variation.
From ANOVA results, it is evident that with fixed input data, EF-LLM produces stable and reliable responses, without generating random or inconsistent outputs.

\section{Future Work}   \label{SEC_future_work}

Although EF-LLM has demonstrated strong AI-assisted automation capability, enhanced sparse prediction performance, and effective hallucination detection ability in the energy forecasting scenarios presented in this paper, it still exhibits certain limitations in terms of scalability and overall capability when facing more complex energy forecasting tasks. This section summarizes the main limitations of EF-LLM at its current stage and, based on the shortcomings of the present work, further outlines the key directions for future research.

\subsection{Continual Learning with Modular LoRA}
At present, the LoRA-based continual learning method illustrated in Fig.\ref{continual_learning_LoRA} is mainly designed for fine-tuning scenarios involving joint time-series and textual inputs. 
Since the prefix block and the main body of the LLM are structurally separated, a newly introduced LoRA module can be used during continual learning to incorporate and update the new knowledge acquired by both components.

  \begin{figure} [!htp]
	\centering
	\includegraphics[width=4in]{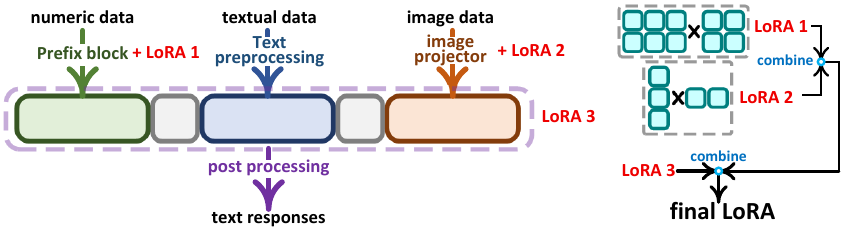}
	\caption{Continual Learning with Modular LoRA.}
	\label{modular_LoRA}
  \end{figure}

However, this method still has limitations when extended to more complex multimodal scenarios. 
As shown in Fig.\ref{modular_LoRA}, future real-world applications may require the fusion of numerical, textual, and visual data for prediction or classification tasks. 
Under such an architecture, the prefix block for processing numerical information and the image projector for processing visual information are usually arranged as parallel modules rather than as a unified sequential structure. 
As a result, LoRA 1 and LoRA 2, which are associated with these modality-specific modules, cannot be directly combined with LoRA 3, which is applied to the main LLM body, in the same straightforward manner as in Fig.\ref{continual_learning_LoRA}. 
This means that the continual learning strategy adopted in Fig.\ref{continual_learning_LoRA} is not directly applicable to this type of multimodal fusion framework.

To address this issue, additional mathematical fusion methods are needed. Specifically, LoRA 1 and LoRA 2 should first be effectively merged, and then further integrated with LoRA 3 to form a final unified LoRA module capable of storing the knowledge of the entire model. 
The method shown in Fig.\ref{modular_LoRA} not only enables the modular LoRA components from different parts of the architecture to be combined into a larger continual-learning-oriented LoRA block, which is not only beneficial for knowledge storage, transfer, and global model updating, but also preserves the possibility of independent fine-tuning for each module. 
Therefore, it offers stronger scalability and greater flexibility.

Nevertheless, the actual performance and effectiveness of this method still require further investigation. More application scenarios that are well aligned with this framework, especially multimodal prediction or classification tasks involving numerical, textual, and visual data, are needed to thoroughly evaluate its feasibility and practical advantages.

\subsection{Limitations of EF-LLM and LLM-MAS-Based Solutions} \label{future_work_limitations}
{It should be clarified that the current function-calling capability of EF-LLM does not originate from the fine-tuning process conducted in this study, but rather from the tool-use capability already provided by its foundation model. This does not imply that such capability cannot be learned through fine-tuning in general.}

{In fact, the authors have publicly released code on IEEE DataPort that can be used to fine-tune various LLMs for tool calling in energy-domain feature engineering tasks \cite{data_2025_zihang}. In that work, 11,100 samples were used to fine-tune LLMs so that they could acquire relatively stable feature-engineering capabilities for energy applications.}

{However, it is important to emphasize that the 11,100-sample dataset for feature-engineering tool-calling and the over 100,000 sequential samples used to train EF-LLM in this study correspond to fundamentally different task levels. The former focuses on structured tool invocation and feature construction behaviors, whereas the latter focuses on time-series forecasting and multimodal representation learning. Therefore, these datasets are not directly comparable in terms of scale or learning objective.}

{In LLM fine-tuning, when heterogeneous objectives are jointly optimized within a single model, relatively small-scale and sparse supervision signals are often dominated by the primary training objective. In contrast, in some real-world applications, certain tool-use capabilities are even learned with only around 100 samples, where the goal is to induce task-specific behavioral patterns rather than generalizable representation learning.}

{Therefore, when knowledge sources are highly heterogeneous and task objectives differ substantially in granularity, directly injecting all capabilities into a single LLM through a unified fine-tuning process may lead to suboptimal utilization of auxiliary supervision signals. In particular, under joint optimization with a dominant forecasting objective, the learning signal for tool-calling behavior may be partially weakened or overshadowed.}

{To address the limitations of a single-LLM design in the complete forecasting pipeline, we further investigate a modular extension of the system. Specifically, we perform targeted modeling and fine-tuning of different functional components and integrate them into a dedicated LLM-based multi-agent system (LLM-MAS) for energy forecasting. This design is not intended to redefine EF-LLM, but rather to serve as a complementary system-level enhancement. In this architecture, different agents are responsible for different functional stages, enabling an explicit decomposition of feature engineering and forecasting responsibilities. For example, the limitations of EF-LLM in data preprocessing and feature engineering can be alleviated by connecting corresponding agent modules through hard-coded prompts, enabling coordinated system-level execution.
The results are provided in Appendix B.}

{Specifically, the feature-engineering agent acts as a compensatory module that enhances the overall pipeline by providing structured feature construction support, even in scenarios where EF-LLM already exhibits partial capability but limited robustness. In the Appendix, we present empirical results demonstrating the effectiveness of this compensation mechanism across several representative forecasting scenarios. The results show that integrating the feature-engineering agent consistently improves the overall performance of the complete forecasting system.}

\section{Conclusion}   \label{SEC_conclusion}
This paper introduces EF-LLM, an LLM designed for energy system forecasting, addressing key challenges in the field. 
First, EF-LLM continuously learns from load, PV, and wind forecasting tasks through a multi-channel architecture and LoRA-based continual learning. 
It integrates task-specific knowledge across data preparation, feature engineering, prompt engineering, multimodal time-series forecasting, and post-forecast grid decision support. 
Combined with function-calling, EF-LLM handles diverse forecasting tasks seamlessly, enabling non-experts to complete workflows effortlessly without professional intervention, lowering entry barriers, and reducing reliance on experienced engineers. 
Second, after F-PEFT, EF-LLM excels in sparse data scenarios, even outperforming state-of-the-art models in PV and load forecasting. 
With the integration of function-calling technology, EF-LLM not only provides accurate forecasting results in sparse data scenarios using its own capabilities but also calls upon the most suitable forecasting models for scenarios where it is less proficient.
Finally, using multi-task learning, semantic cosine similarity analysis, and ANOVA, EF-LLM is the first energy-focused LLM to identify and quantify hallucination rates, further enhancing its reliability.

In conclusion, EF-LLM is the first energy forecasting LLM capable of achieving AI-assisted automation through human-AI interaction, showing excellent performance in energy forecasting related tasks.
Its flexible knowledge injection framework opens up numerous possibilities for future developments. 
With continuous new data injection, EF-LLM has the potential to integrate a vast array of algorithms, evolving into a fully automated end-to-end energy LLM in the future.

\textbf{Page 29, }

\appendix
\section{How To Use EF-LLM To Complete A Whole Prediction Task} 
\label{Appen_1}

\begin{figure} [!htp]
	\centering
	\includegraphics[width=5.5in]{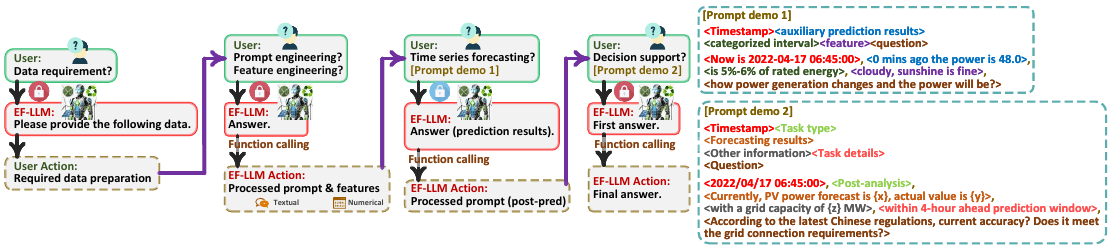}
	\caption{Complete Prediction Task Implementation Using EF-LLM.}
	\label{EF_LLM_reasoning_chain}
  \end{figure}

Fig.\ref{EF_LLM_reasoning_chain} illustrates how EF-LLM can be used to complete a full prediction task. 
For specific time-series tasks, EF-LLM can accomplish the prediction independently without invoking any functions due to its inherent forecasting capabilities. 

\begin{figure} [!htp]
	\centering
	\includegraphics[width=5.5in]{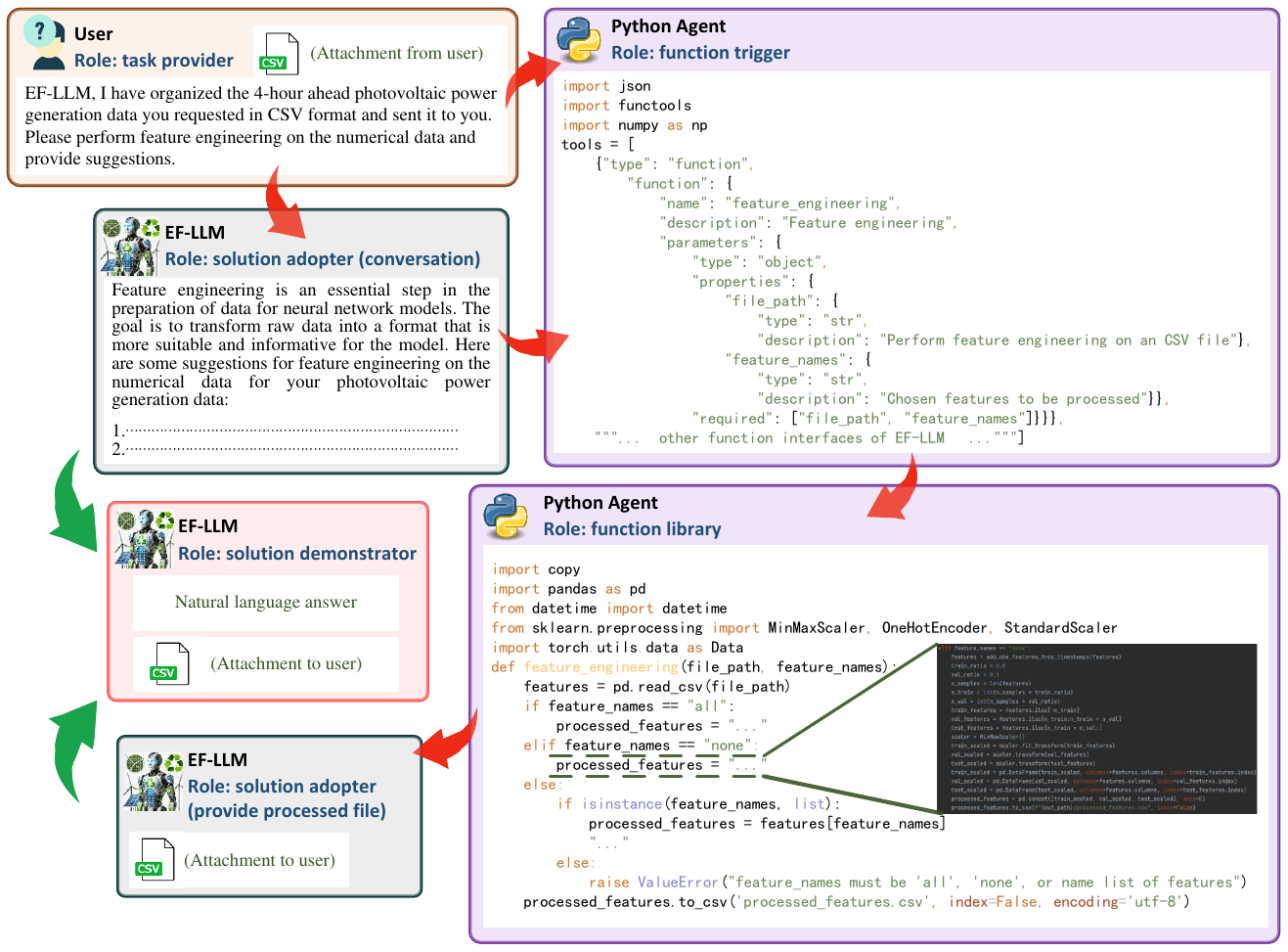}
	\caption{How EF-LLM Completes a Task via Function Calling.}
	\label{function_task}
  \end{figure}

For other tasks, EF-LLM determines the appropriate action based on the user prompt and its initial response, triggering the corresponding function as needed. 
Fig.\ref{function_task} demonstrates an example of how EF-LLM activates the feature engineering function.

\clearpage

Based on Fig.\ref{EF_LLM_reasoning_chain}, the entire process of completing the prediction task is as follows:

\begin{figure}[!htbp]
    \centering
    \includegraphics[width=5in]{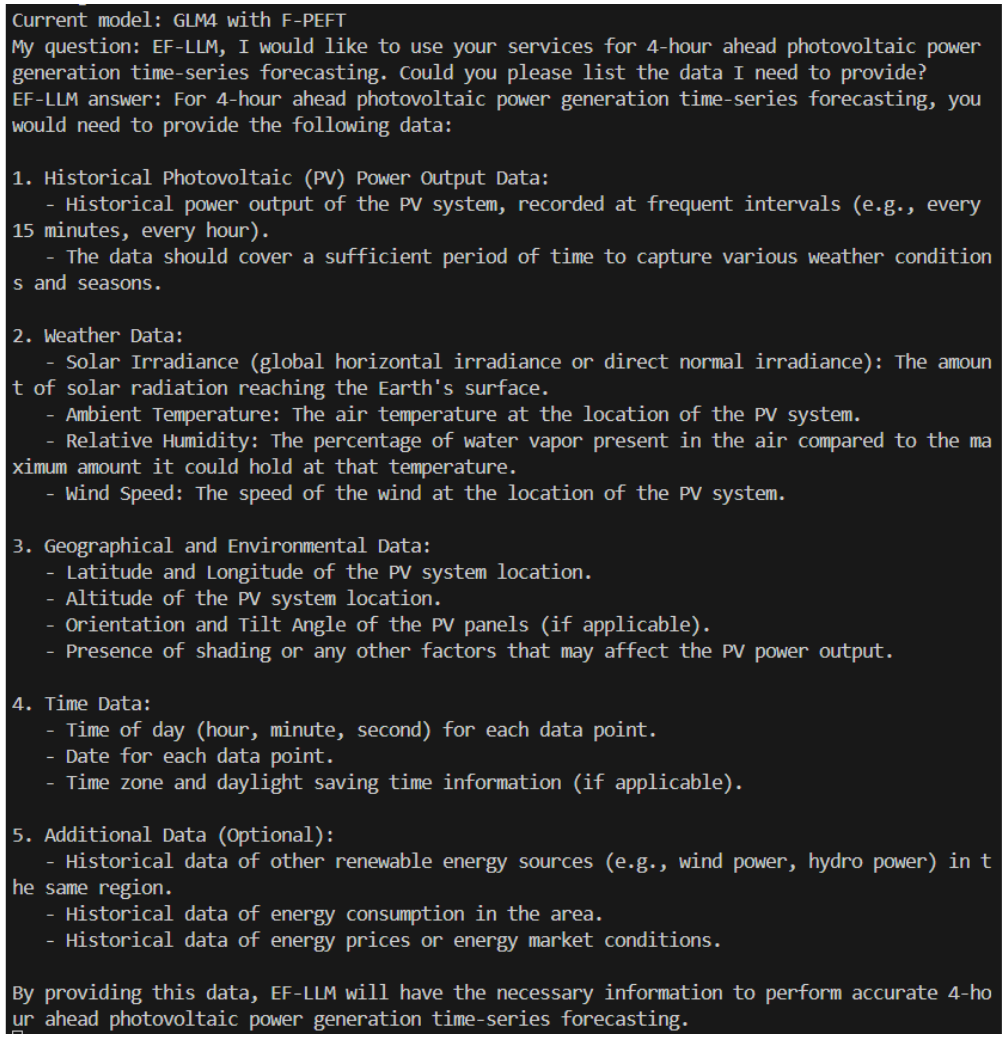}
    \caption{EF-LLM Interaction With User -- Data Preparation.}
    \label{Appendix_1}
\end{figure}
\clearpage

STEP 1: Data Preparation, as in Fig.\ref{Appendix_1}: 
As a case study, information about the feature data required for current PV power prediction task was injected, enabling EF-LLM to more consistently output the features used in the feature engineering process. 
Users can consult EF-LLM about the data they need to provide, receive the corresponding response, and then prepare the required data, organizing it into a table to be provided to EF-LLM.

\begin{figure}[!htbp]
    \centering
    \includegraphics[width=5in]{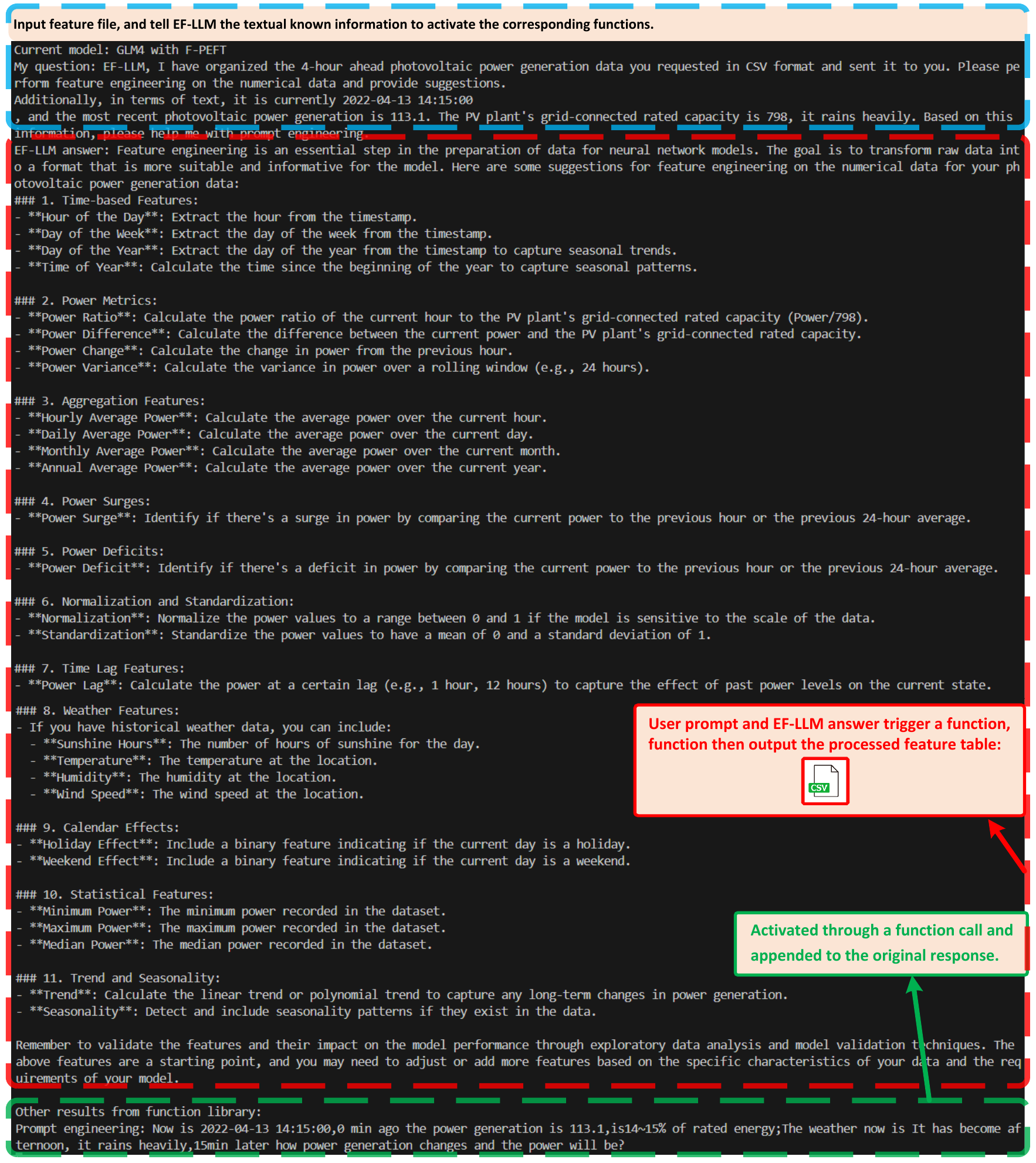}
    \caption{EF-LLM Interaction With User -- Feature \& Prompt Engineering.}
    \label{Appendix_2}
\end{figure}
\clearpage

STEP 2: Feature Engineering \& Prompt Engineering, Fig.\ref{Appendix_2}: 
EF-LLM can process multimodal data through its multi-channel structure.
By interacting with EF-LLM and providing prepared numerical data tables, users can activate EF-LLM to call corresponding functions from its function library. 
Users may choose to inquire about feature engineering, prompt engineering, or both, depending on the task at hand.
1.For pure prediction tasks in STEP 3, EF-LLM requires both numerical data and a prompt. 
In this case, as illustrated in the example in Fig.\ref{Appendix_2}, users need to provide both the numerical table and prompt to EF-LLM. 
EF-LLM will call the respective functions and return the processed features and prompt.
2.For incomplete feature tasks, where the prediction is already completed but additional features need to be supplemented (and these features were described during the F-PEFT process), users only need to inquire about prompt engineering.
3.For training tasks using models other than EF-LLM, users can request EF-LLM to handle feature engineering only. 
EF-LLM will process the input table, respond accordingly, and provide the required feature table through function calling.

\begin{figure}[!htbp]
    \centering
    \includegraphics[width=5in]{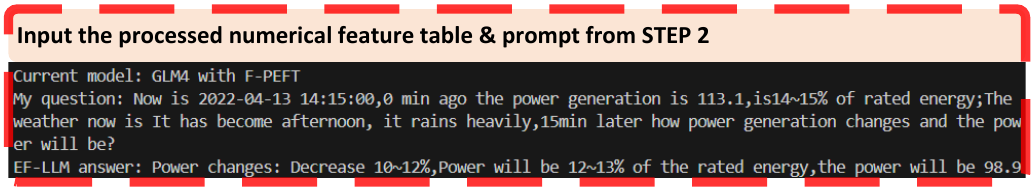}
    \caption{EF-LLM Interaction with User -- Multimodal Time Series Prediction.}
    \label{Appendix_3}
    \vspace{1em}
    \includegraphics[width=5in]{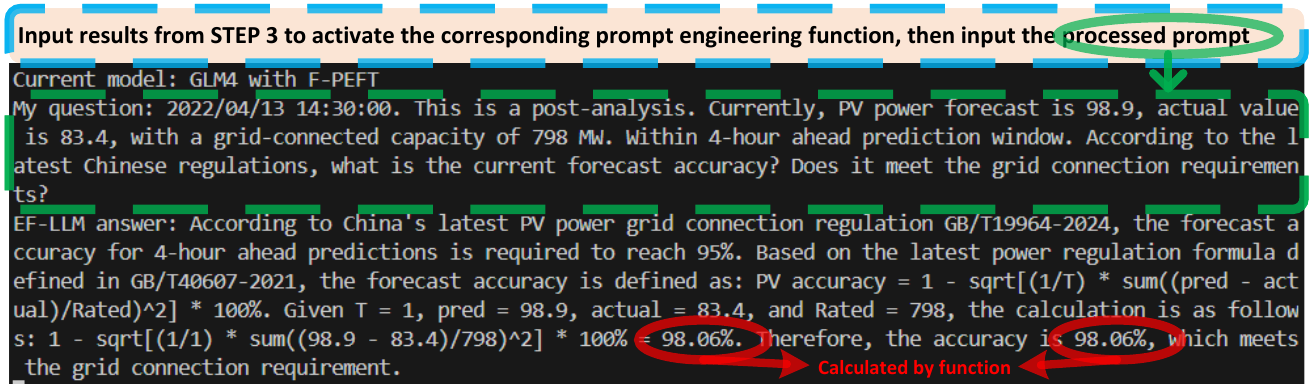}
    \caption{EF-LLM Interaction with User -- Post-prediction Decision Support.}
    \label{Appendix_4}
\end{figure}

STEP 3: Multimodal Time-Series Prediction, Fig.\ref{Appendix_3}. 
After obtaining the numerical feature table in STEP 2, the numerical feature table and the processed prompt are input into EF-LLM. 
EF-LLM will then directly provide the prediction results.

STEP 4: Post-Prediction Decision Support, Fig.\ref{Appendix_4}. 
EF-LLM utilizes the prediction results from STEP 3 and the user’s specific requirements to call the corresponding functions. 
It then provides decision support, completes the task, and delivers the results and conclusions.

\section{{Effect of EF-LLM after compensation by the feature engineering module}} 
\label{Appen_FE_extension}
\begin{figure}[!htbp]
    \centering
    \includegraphics[width=5in]{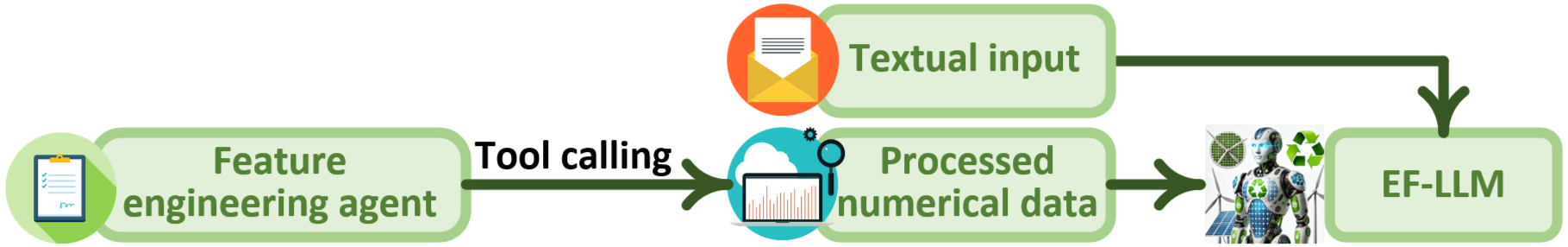}
    \caption{{Feature Engineering Module and EF-LLM Compensation Framework.}}
    \label{Appendix_5}
\end{figure}

{As discussed in Section \ref{future_work_limitations}, EF-LLM has relatively limited capability in calling tools for the feature engineering module, and it is difficult for EF-LLM to accurately invoke complex feature engineering tools. This limitation is mainly caused by the large differences in task data scale during LLM fine-tuning and continual learning, making it difficult to fully develop such capabilities within a single LLM. Therefore, we designed the compensation mode shown in Fig. \ref{Appendix_5}. Specifically, an energy-domain feature engineering agent is fine-tuned to perform refined feature engineering on the numerical data. The processed data are then fed into the Multi-Channel Architecture of EF-LLM together with textual information, enabling interactive forecasting.}

{Here, the feature engineering agent module used to compensate for EF-LLM’s feature engineering capability in energy forecasting has undergone a specialized two-stage targeted fine-tuning process, namely Supervised Fine-Tuning and Group Relative Policy Optimization (SFT-GRPO). 
Its fine-tuning process, code, and data have all been publicly released \cite{qiu2026energyfetool}.}

\begin{table} [!htb] 
  	\centering
  	\caption{{Prediction Performance Under Feature Sparsity After Feature Engineering Module Compensation}}
        \label{TB_appen_1}
	\renewcommand\arraystretch{1.5}
  	\begin{tabular} {c|c|c|c|c|c|c}
  		\hline
  		\hline
  		Load  &  EF-CLS   &  EF-REG  &  LSTM  &  iTF  &  TN  &  STELLM-style  \\   
  		\hline
  		MAE (kW) &   17.11   &   24.36   &   20.82   &   17.65    &  18.10     &   18.23 \\ 
  		\hline
  		RMSE (kW) &  17.91   &   33.86   &   25.47   &   17.33    &  20.67     &   20.79 \\ 
  		\hline
  		\hline
  		PV &  EF-CLS   &  EF-REG  &  LSTM  &  iTF  & TN   &   STELLM-style  \\   
  		\hline
  		MAE (kW) &    14.81   &   17.16   &   15.98   &    14.35   & 14.04      &   15.39 \\  
  		\hline
  		RMSE (kW) &    36.75   &   43.87   &   43.24   &    33.94   &   34.24    &   44.78   \\ 
  		\hline
  		\hline
  		Wind  &  \makecell{EF-CLS}   &  \makecell{EF-REG}  &  LSTM  &  iTF  & TN   &   STELLM-style  \\   
  		\hline
  		MAE (W) &    93.54   &   128.74   &   95.32  &    85.25   &  79.28     &   92.72 \\ 
  		\hline
  		RMSE (W) &   176.64    &   243.10   &   185.25   &   169.65    &   146.75    &   178.10 \\ 
            \hline
            \hline
  	\end{tabular}  
  \end{table}

{From Table \ref{TB_appen_1}, it can be observed that although the feature-engineering tool-calling capability of EF-LLM itself is subject to certain limitations, this issue can be effectively addressed by integrating it with a separately fine-tuned energy-domain feature engineering agent, thereby further improving the model’s prediction accuracy.}

\section*{Acknowledgements}
The authors gratefully acknowledge Assoc. Prof. Guo Chen and Prof. Chaojie Li from UNSW Sydney for their valuable suggestions on this work.

\begin{footnotesize}
\bibliographystyle{elsarticle-num}
\bibliography{cite}
\end{footnotesize}

\end{document}